\documentclass[10pt,journal,compsoc]{IEEEtran}

\ifCLASSOPTIONcompsoc
  \usepackage[nocompress]{cite}
\else
  \usepackage{cite}
\fi

\usepackage{graphicx}
\usepackage{subcaption}
\usepackage{multirow}
\usepackage{amssymb}
\usepackage{amsmath}
\usepackage{booktabs}
\usepackage{algorithm}
\usepackage{algorithmic}

\usepackage{color}
\usepackage{ragged2e}

\usepackage{hyperref}

\usepackage{bbding}
\usepackage{color}

\newcommand{\tabincell}[2]{\begin{tabular}{@{}#1@{}}#2\end{tabular}}

\ifCLASSINFOpdf

\else

\fi

\begin{document}

\title{SiMaN: Sign-to-Magnitude Network Binarization}

\author{Mingbao Lin,
        Rongrong Ji,~\IEEEmembership{Senior Member,~IEEE},
        Zihan Xu,
        Baochang Zhang,~\IEEEmembership{Senior Member,~IEEE},\\
        Fei Chao,~\IEEEmembership{Member,~IEEE},
        Chia-Wen Lin~\IEEEmembership{Fellow,~IEEE}
        and Ling Shao,~\IEEEmembership{Fellow,~IEEE}
\IEEEcompsocitemizethanks{
\IEEEcompsocthanksitem M. Lin, R. Ji (Corresponding Author), Z. Xu and F. Chao are with the Media Analytics and Computing Laboratory, Department of Artificial Intelligence, School of Informatics, Xiamen University, Xiamen 361005, China (e-mail: rrji@xmu.edu.cn).\protect
\IEEEcompsocthanksitem M. Lin and Z. Xu are also with the Tencent Youtu Lab, Shanghai 200233, China.\protect
\IEEEcompsocthanksitem B. Zhang is with the Zhongguancun Lab, Beijing 100190, China.\protect
\IEEEcompsocthanksitem C.-W. Lin is with the Department of Electrical Engineering and the Institute of Communications Engineering, National Tsing Hua University, Hsinchu 30013, Taiwan.
\IEEEcompsocthanksitem L. Shao is with Terminus Group, China.}
\thanks{Manuscript received April 19, 2005; revised August 26, 2015.}}

\markboth{IEEE TRANSACTIONS ON PATTERN ANALYSIS AND MACHINE INTELLIGENCE UNDER REVIEW}%
{Shell \MakeLowercase{\textit{et al.}}: Bare Demo of IEEEtran.cls for Computer Society Journals}

\IEEEtitleabstractindextext{%
\begin{abstract}
\justifying
Binary neural networks (BNNs) have attracted broad research interest due to their efficient storage and computational ability. Nevertheless, a significant challenge of BNNs lies in handling discrete constraints while ensuring bit entropy maximization, which typically makes their weight optimization very difficult. Existing methods relax the learning using the sign function, which simply encodes positive weights into $+1$s, and $-1$s otherwise. Alternatively, we formulate an angle alignment objective to constrain the weight binarization to $\{0,+1\}$ to solve the challenge. In this paper, we show that our weight binarization provides an analytical solution by encoding high-magnitude weights into $+1$s, and $0$s otherwise. Therefore, a high-quality discrete solution is established in a computationally efficient manner without the sign function. We prove that the learned weights of binarized networks roughly follow a Laplacian distribution that does not allow entropy maximization, and further demonstrate that it can be effectively solved by simply removing the $\ell_2$ regularization during network training. Our method, dubbed sign-to-magnitude network binarization (SiMaN), is evaluated on CIFAR-10 and ImageNet, demonstrating its superiority over the sign-based state-of-the-arts. Our source code, experimental settings, training logs and binary models are available at \url{https://github.com/lmbxmu/SiMaN}.
\end{abstract}

}

\maketitle

\IEEEdisplaynontitleabstractindextext

\IEEEpeerreviewmaketitle

\section{Introduction \label{introduction}}

\IEEEPARstart{D}{eep} neural networks (DNNs), especially convolutional neural networks (CNNs), have been effectively used in many tasks of computer vision, such as image recognition~\cite{he2016deep,wei2015hcp,he2015spatial}, object detection~\cite{redmon2016you,ren2016object,zhang2021learning}, and semantic segmentation~\cite{long2015fully,badrinarayanan2017segnet,shelhamer2017fully}. Nowadays, DNNs are almost trained on high-capacity but power-hungry graphics processing units (GPUs); however, such DNN models often fail to run on low-power devices such as cell phones and Internet-of-Things (IoT) devices that have been universally popularized in modern society. As a result, substantial efforts have been invested to reduce the model redundancy while retaining a comparable or even better accuracy performance in comparison with the full model, such that the compressed model can be easily deployed on these resource-limited devices.

Typical methods for reducing the model redundancy include, but are not limited to:
(1) Weight pruning discards individual weights in the filters or connections across different layers, and then reshapes the model in a sparse format~\cite{han2015learning,frankle2019lottery}.
(2) Filter pruning resorts to directly removing all weights in a filter and the corresponding channel in the next layer~\cite{luo2018thinet,lin2020hrank}.
(3) Compact network designs, such as ShuffleNets~\cite{zhang2018shufflenet,ma2018shufflenet}, MobileNets~\cite{howard2017mobilenets,sandler2018mobilenetv2,howard2019searching} and GhostNet~\cite{han2020ghostnet}, choose to directly build parameter-efficient neural network models.
(4) Tensor decomposition approximates the weight tensor with a series of low-rank matrices, which are then reorganized in a sum-product form~\cite{lin2018holistic,hayashi2019exploring} to recover the original weight tensor.
(5) Low-precision quantization aims to compress the model by reducing the number of bits used to represent the weight parameters of the pre-trained models~\cite{cai2017deep,han2020training,lin2020rotated}.

In particular, binary neural networks (BNNs), which quantize their weights and activations in a 1-bit binary form, have attracted increasing attention for two major reasons: 1) The memory usage of a BNN is 32$\times$ lower than its full-precision counterpart, since the weights of the latter are stored in a 32-bit floating-point form. 2) A significant reduction of computational complexity can be achieved by executing efficient XNOR and bitcount operations, \emph{e.g.}, up to 58$\times$ speed-ups on CPUs as reported by~\cite{rastegari2016xnor}. Regardless of these two merits, BNNs are also famed for their significant performance degradation. For example, XNOR-Net~\cite{rastegari2016xnor} suffers an $~$18\% drop in top-1 accuracy when binarizing ResNet-18 on the ImageNet classification task~\cite{deng2009imagenet}. The poor performance greatly barricades the possibility of deploying BNNs in real-world applications.

One of the major obstacles in constructing a high-performing BNN is the discrete constraints imposed on the pursued binary weights, which challenges the weight optimization. Meanwhile, BNNs also require the two possible values of binarized weights to be uniformly (half-half) distributed to ensure bit entropy maximization. To this end, most existing approaches simply employ the sign function to binarize weights where positive weights are encoded into $+1$s, and $-1$s are used otherwise~\cite{rastegari2016xnor,lin2017towards,liu2018bi,qin2020forward,lin2020rotated}. To compensate for the entropy information, recent methods, such as Bayesian optimization~\cite{gu2019bayesian}, rotation matrix~\cite{lin2020rotated}, and weight standardization~\cite{qin2020forward}, learn a two-mode distribution for real-valued weights to increase the probability of encoding one half of the weights  into $+1$s and the other half into $-1$s by the sign function. These strategies, however, increase the learning complexity, since the optimization involves additional training loss terms and variables. Moreover, it is unclear whether the simple usage of the sign function is the optimal encoding option for the weight binarization process.

\begin{figure}
\begin{center}
\includegraphics[height=0.46\linewidth]{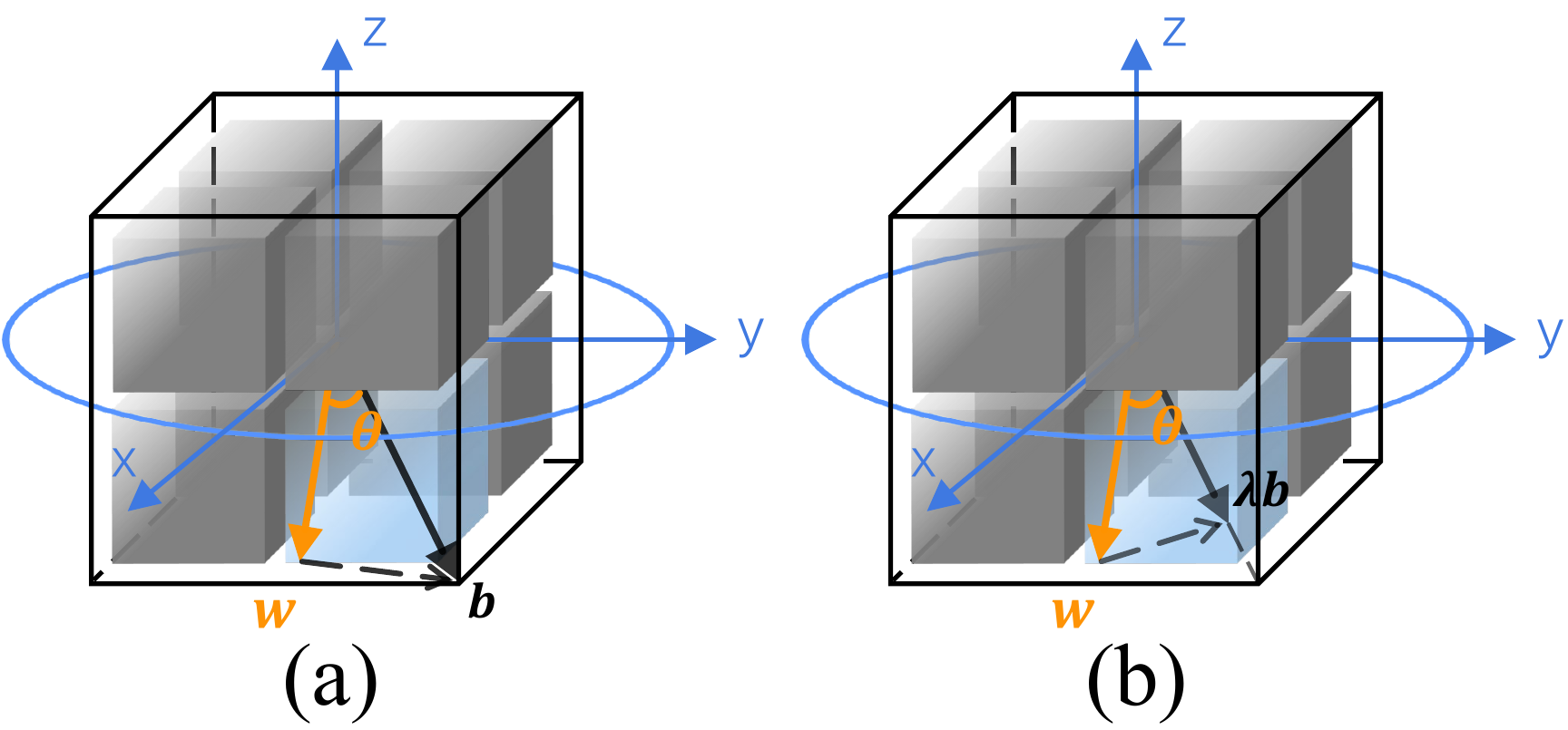}
\end{center}
\vspace{-0.8em}
\caption{\label{error} (a) Early works \cite{courbariaux2015binaryconnect,courbariaux2016binarized} suffer from large quantization error caused by both the norm gap and angular bias between the full-precision weights and its binarized version. (b) Recent works \cite{rastegari2016xnor,qin2020forward} introduce a scaling factor to reduce the norm gap but cannot reduce the angular bias, \emph{i.e.}, $\theta$. Therefore the quantization error $\|  \mathbf{w}\sin\theta \|^2$ is still large when $\theta$ is large.
\vspace{-1.0em}
}
\end{figure}

Another obstacle in learning BNNs comes at the large quantization error between the full-precision weight vector $\mathbf{w}$ and its binary vector $\mathbf{b}$ \cite{courbariaux2015binaryconnect,courbariaux2016binarized} as illustrated in Fig.\,\ref{error}(a). To solve this, state-of-the-art approaches \cite{rastegari2016xnor,qin2020forward} introduce a per-channel learnable/optimizable scaling factor $\lambda$ to decrease the quantization error
\begin{equation}\label{scaling}
\mathop{\min}_{\lambda, \mathbf{b}} \| \lambda\mathbf{b} - \mathbf{w} \|^2.
\end{equation}
However, as revealed in the earlier version of this paper~\cite{lin2020rotated}, the introduction of $\lambda$ only partly mitigates the quantization error by compensating for the norm gap between the full-precision weight and its binarized version, but cannot reduce the quantization error due to an angular bias as shown in Fig.\,\ref{error}(b). Apparently, with a fixed angular bias $\theta$, when $\lambda\mathbf{b}-\mathbf{w}$ is orthogonal to $\lambda\mathbf{b}$, Eq.\,(\ref{scaling}) reaches the minimum and we have
\begin{equation}
\| \mathbf{w}\sin\theta \|^2 \le \| \lambda\mathbf{b} - \mathbf{w} \|^2.
\end{equation}

\begin{figure}[!t]
\begin{center}
\includegraphics[width=0.8\linewidth]{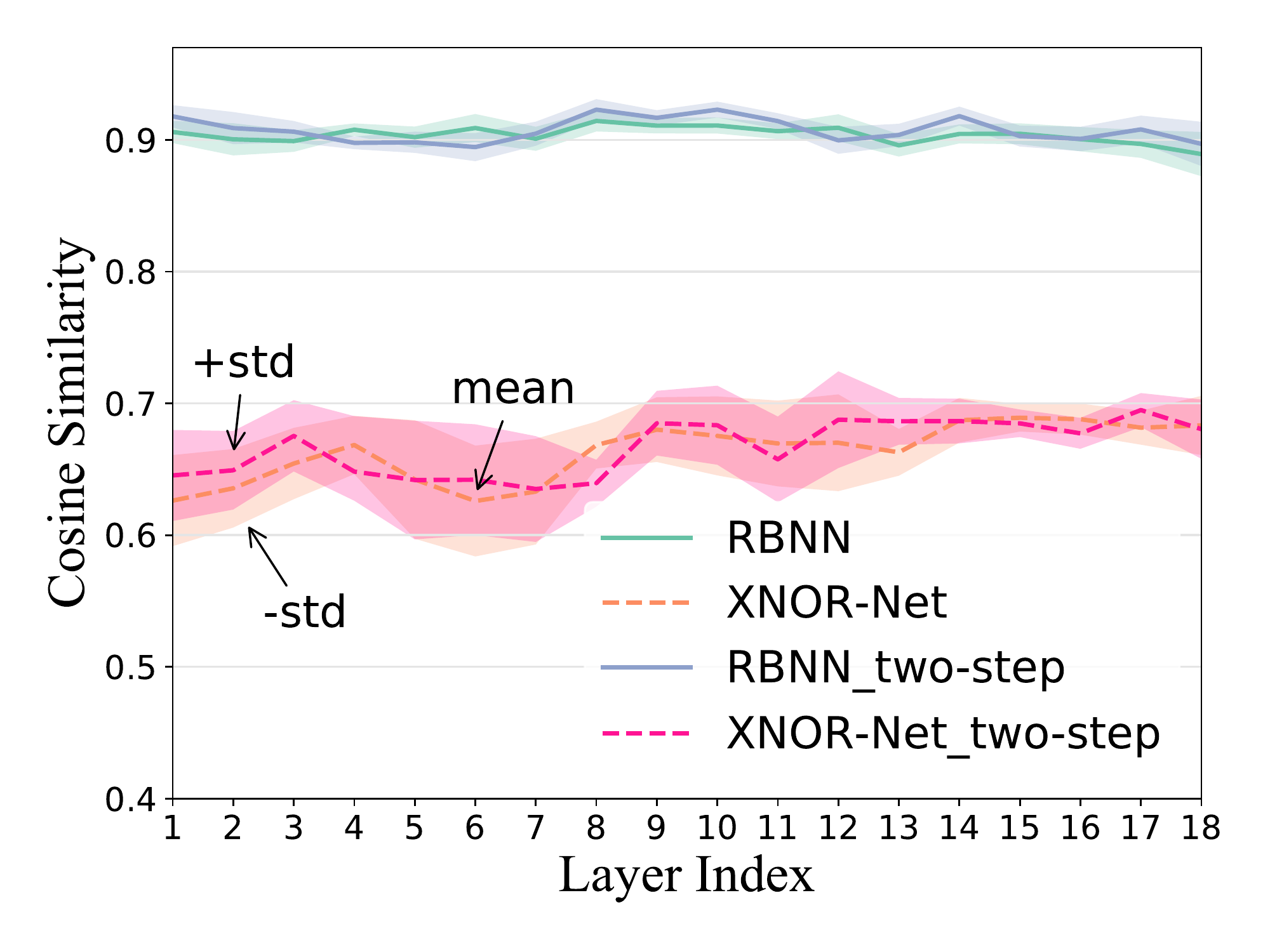}
\end{center}
\vspace{-0.8em}
   \caption{Cosine similarity between the full-precision weight vector and the corresponding binary vector in various layers of ResNet-20.}
\vspace{-1.0em}
\label{cosine-similarity}
\end{figure}

Thus, the $\| \mathbf{w}\sin\theta \|^2$ serves as the lower bound of the quantization error and cannot be diminished as long as the angular bias exists. This lower bound could be huge with a large angular bias $\theta$. 
Though the training process updates the weights and may close the angular bias, we experimentally observe the possibility of this case is small, as illustrated by XNOR-Net~\cite{rastegari2016xnor} in Fig.\,\ref{cosine-similarity}.
Thus, it is natural for researchers to go further reduce this angular error for the sake of minimizing the quantization error if a better BNN performance is anticipated to obtain.

\begin{figure}[!t]
\begin{center}
\includegraphics[height=0.45\linewidth]{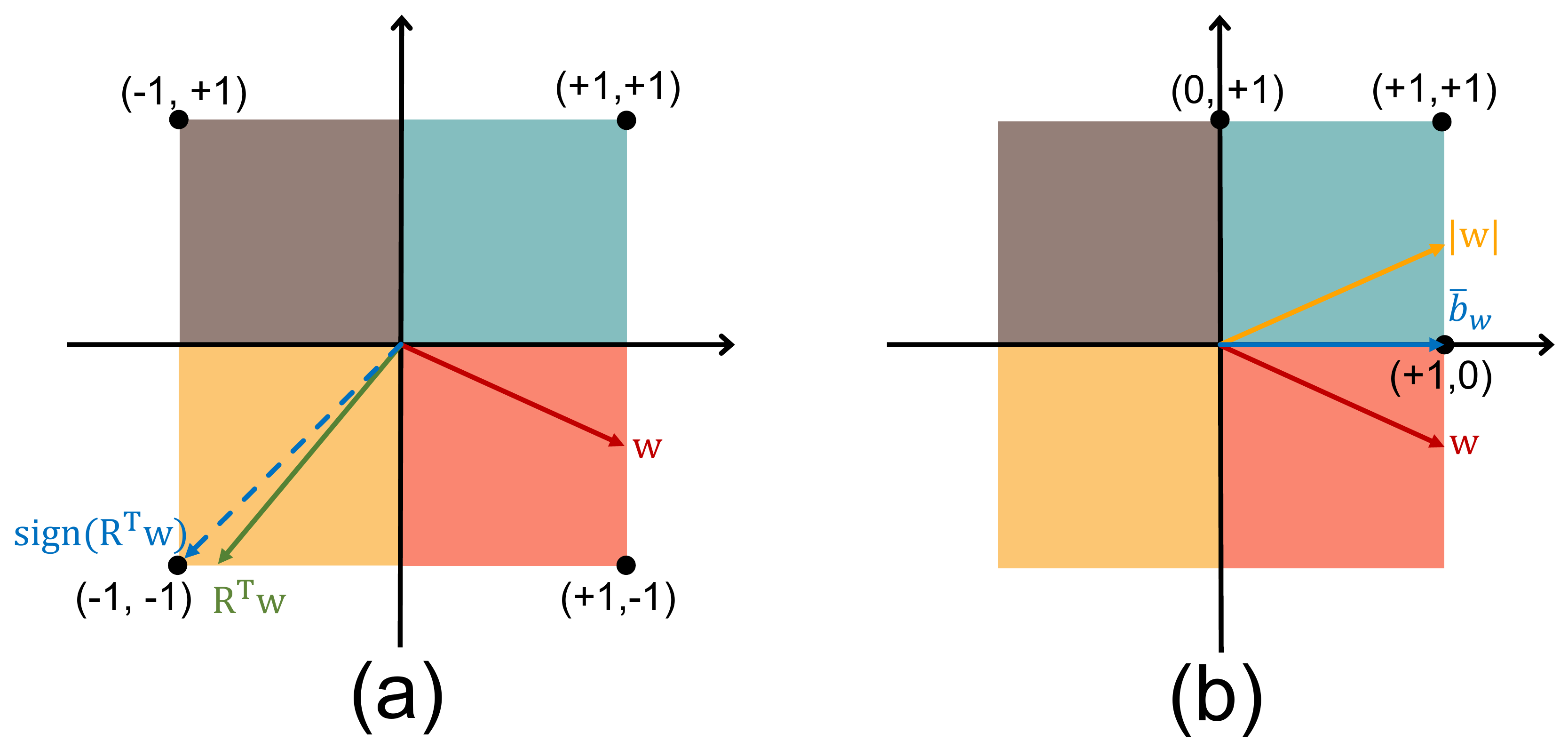}
\end{center}
\vspace{-0.8em}
\caption{\label{bnn}
Comparison between (a) the preliminary version of RBNN~\cite{lin2020rotated} and (b) the extended version of SiMaN in this paper. RBNN learns a rotation matrix $\mathbf{R}$ first, and then applies the sign function to binarize the rotated weight $\mathbf{b}_w = \operatorname{sign}(\mathbf{R}^T\mathbf{w}) \in \{-1, +1\}^n$. In contrast, the presented SiMaN in this paper involves the magnitude of the weight, and then discretely learns $\bar{\mathbf{b}}_w \in \{0, 1\}^n$.
}
\vspace{-1.0em}
\end{figure}

To solve the angular bias, the earlier version~\cite{lin2020rotated} proposed the angle alignment based learning objective which is originally formulated as
\begin{equation}\label{rotation}
\begin{split}
  \mathop{\arg\max}_{\mathbf{R}} & \frac{\operatorname{sign}(\mathbf{R}^T\mathbf{w})^T (\mathbf{R}^T\mathbf{w})}{\| \operatorname{sign}(\mathbf{R}^T\mathbf{w}) \|_2 \| \mathbf{R}^T\mathbf{w} \|_2}, 
  \\&
   \; s.t. \quad \mathbf{R}^T\mathbf{R} = \mathbf{I}_n,
\end{split}
\end{equation}
where $\mathbf{R}$ is constrained to an $n$-order rotation matrix. As shown in Fig.\,\ref{bnn}(a), by applying the sign function on the rotated weight vector $\mathbf{R}^T\mathbf{w}$, we attain the binarization of $\mathbf{w}$, \emph{i.e.}, $\mathbf{b}_w \in \operatorname{sign}(\mathbf{R}^T\mathbf{w})$. Thus, Eq.\,(\ref{rotation}) aims to learn a rotation matrix such that the angle bias between the rotated weight vector and its encoded binarization is reduced as illustrated by RBNN~\cite{lin2020rotated} (conference version) in Fig.\,\ref{cosine-similarity}.
Though the great reduction on quantization error has been quantitatively measured in~\cite{lin2020rotated}, the learning complexity of the rotation matrix, $\mathbf{R}$, is very high, due to the non-convexity of Eq.\,(\ref{rotation}). Thus, an alternating optimization approach is developed. Nevertheless, the alternating optimization results in sub-optimal binarization. Moreover, the optimization is still built upon the sign function.
Note that, in Fig.\,\ref{cosine-similarity}, we also train XNOR-Net and RBNN with the two-step training paradigm~\cite{martinez2019training}. We can see that the angular bias remains similar to these with commonly-used from-scratch training. Thus, training BNNs with different strategies does not correct the angular bias.

In this paper, a novel sign-to-magnitude network binarization (SiMaN) is proposed to discretely encode DNNs, leading to improved accuracy. Within our method, we reformulate the angle alignment objective in the conference version~\cite{lin2020rotated}, which aims to maximize the cosine distance between the full-precision weight vector and its encoded binarization. Different from existing works that binarize weights into $\{-1, +1\}$ by the sign function, our binarization falls into $\{0, +1\}$ as illustrated in Fig.\,\ref{bnn}(b). In this way, we reveal that the globally analytical binarization for our angle alignment can be found in a computationally efficient manner of $\mathcal{O}(n\log n)$ by quantizing into $+1$s the high-magnitude weights, and $0$s otherwise, therefore enabling weight binarization without the sign function. To the best of our knowledge, we prove for the first time that the learned real-valued weights roughly follow a Laplacian distribution, which results in around 37\% of weights being encoded into $+1$s. This prevents the BNN from maximizing the entropy of information. To solve this, we do not add a term to the loss function since this increases the optimization difficulty. Alternatively, we analyze the intrinsic numerical values of weights, and show that the simple removal of the $\ell_2$ regularization destroys the Laplacian distribution, and thus enhances the half-half weight binarization. As a result, the final binarization is obtained by encoding into $+1$s weights within the largest top-half magnitude, and $0$s otherwise to further reduce the computational complexity from $\mathcal{O}(n \log n)$ to $\mathcal{O}(n)$.

A preliminary conference version of this work was presented in~\cite{lin2020rotated}. The main contributions we have made in this paper are listed in the following.

\begin{itemize}
 \item A new learning objective based on the angle alignment is proposed and a magnitude-based analytical solution for BNNs is developed in a computationally efficient manner.
 \item We formally prove that the learned weights in BNNs follow a Laplacian distribution, which, as revealed, prevents the maximization of bit entropy.
 \item A detailed analysis on the numerical values of weights shows that simply removing the $\ell_2$ regularization benefits maximizing the bit entropy while further reducing the computational complexity.
 \item Experiments on CIFAR-10~\cite{krizhevsky2009learning} and ImageNet~\cite{deng2009imagenet} demonstrate that our sign-to-magnitude framework for network binarization outperforms the traditional sign-based binarization.
\end{itemize}

\section{Related Work \label{related_work}}

Following the introduction of pioneering research~\cite{courbariaux2016binarized} where the sign function and the straight-through estimator (STE)~\cite{bengio2013estimating} are respectively adopted for the forward weight/activation binarization and backward gradient updating, BNNs have emerged as one of the most appealing approaches for the deployment of DNNs in resource-limited devices. As such, great efforts have been put into closing the gap between full-precision networks and their BNNs. In what follows, we briefly review some related works. A comprehensive overview can be found in the survey papers~\cite{simons2019review,qin2020binary}.

XNOR-Net~\cite{rastegari2016xnor} introduces two scaling factors for channel-wise weights and activations to minimize quantization error. Inspired by this, XNOR-Net++~\cite{bulat2019xnor} improves the performance by integrating the two scaling factors into one, which is then updated using the standard gradient propagation. Except for the scaling factors, RBNN~\cite{lin2020rotated} further reduces the quantization error by optimizing the angle difference between a full-precision weight vector and its binarization. 
Xu \emph{et al}.~\cite{xu2021recu} observed ``dead weights'' in binary neural networks and proposed to mitigate quantization error by clipping large-magnitude weights to a fixed element.
To enable the gradient propagation and reduce the ``gradient mismatch'' by the STE~\cite{bengio2013estimating}, several works,  such as the swish function~\cite{darabi2018bnn+}, piece-wise polynomial function~\cite{liu2018bi}, and error decay estimator~\cite{qin2020forward}, formulate the forward/backward quantization as a differentiable non-linear mapping. 
FDA~\cite{xu2021learning} estimates the gradient of sign function in the Fourier frequency domain using the combination of sine function for training BNNs.
Another direction circumvents the gradient approximation of the sign function by sampling from the weight distribution~\cite{peters2018probabilistic,shayer2018learning}. 
Qin \emph{et al}.~\cite{qin2020bipointnet} introduces entropy-maximizing aggregation to modulate the distribution for the maximum information entroy, and layer-wise scale recovery to restore feature representation capacity.
There are also abundant works that explore the optimization of BNNs~\cite{leng2018extremely,alizadeh2018empirical,bethge2019back,helwegen2019latent,martinez2019training} and explain their effectiveness~\cite{anderson2018high}. 
Wang \emph{et al}.~\cite{wang2021sub} proposed to train BNNs under a kernel-aware optimization framework.
ProxConnect (PC)~\cite{dockhorn2021demystifying} generalizes and improves BinaryConnect (BC) with well-established theory and algorithms.
Recent works~\cite{darabi2018bnn+,gu2019projection} embed various regularization terms into the training loss to binarize the weights and control the activation ranges~\cite{ding2019regularizing}. 
Hu \emph{et al}.~\cite{hu2022elastic} added real-valued input features to the subsequent convolutional output features to enrich information flow within a BNN.
Moreover, other recent studies devise binarization-friendly structures to boost the performance. For example, Bi-Real~\cite{liu2018bi} designs double residual connections with full-precision downsampling layers. XNOR-Net++~\cite{bulat2019xnor} replaces ReLU by PReLU. ReActNet~\cite{liu2020reactnet} adds parameter-free shortcuts on MobileNetV1~\cite{howard2017mobilenets} and the group convolution is replaced by a regular convolution.

\section{Binary Neural Networks \label{binary_neural_networks}}

For an $L$-layer CNN model, we denote $\mathbf{W}^i = \{ \mathbf{w}^i_1, \mathbf{w}^i_2, ..., \mathbf{w}^i_{c^i_{out}} \} \in \mathbb{R}^{n^i \times c^i_{out} }$ as the real-valued weight set for the $i$-th layer, where $\mathbf{w}^i_j \in \mathbb{R}^{n^i}$ denotes the $j$-th weight. The real-valued input activations of the $i$-th layer are represented as $\mathbf{A}^i = \{ \mathbf{a}^i_1, \mathbf{a}^i_2, ..., \mathbf{a}^i_{c^i_{in}} \} \in \mathbb{R}^{m^i \times c^i_{in}}$; here, $c^i_{out}$ and $c^i_{in}$ respectively represent the output and input channels, and $n^i$ and $m^i$ denote the size of each weight and input, respectively. Then, the convolution result can be expressed by
\begin{equation}\label{full_conv}
 \mathbf{a}^{i+1}_j = \mathbf{w}^i_j \circledast \mathbf{A}^i,
\end{equation}
where $\circledast$ stands for the convolution operation. For simplicity, we omit the non-linear layer here. 

\textbf{BNN Training}.
To train a BNN, the real-valued $\mathbf{w}^i_j$ and $\mathbf{A}^i$ in Eq.\,(\ref{full_conv}) are quantized into binary values  $(\mathbf{b}_w)^i_j \in \{-1, +1\}^{n^i}$ and $(\mathbf{B}_A)^i \in \{-1, +1\}^{c^i_{in} \times m^i}$, respectively. As a result, the convolution result can be approximated as
\begin{equation}\label{approximated_convolution}
 \mathbf{a}^{i+1}_j \approx {\beta}^i_j \cdot (\mathbf{b}_w)_j^i \circledast (\mathbf{B}_A)^i,
\end{equation}
where ${\beta}^i_j$ is a channel-level scaling factor~\cite{rastegari2016xnor,bulat2019xnor}.

For the implementation of the BNN training, the forward calculation is fulfilled by conducting the convolution between $(\mathbf{b}_w)_j^i$ and $(\mathbf{B}_A)^i$ in Eq.\,(\ref{approximated_convolution}), whereas their real-valued counterparts, $\mathbf{w}_j^i$ and $\mathbf{A}^i$, are updated during backpropagation. To this end, following existing studies~\cite{hubara2016binarized,bulat2019xnor,lin2020rotated}, the activation binarization in this work is simply realized by the sign function as
\begin{equation}\label{binarization}
    (\mathbf{B}_A)^i = \operatorname{sign}(\mathbf{A}^i)=\left\{\begin{array}{l}
        +1, \text{ if } \mathbf{A}^i \ge 0, \\
        -1, \text{ otherwise.}
    \end{array}\right. 
\end{equation}

In the backpropagation phase, we adopt the piece-wise polynomial function~\cite{liu2018bi} to approximate the gradient of a given loss $\mathcal{L}$ \emph{w.r.t}. the input activations $\mathbf{A}^i$  as follows
\begin{equation}\label{gradient_activation}
  \frac{\partial \mathcal{L}}{\partial \mathbf{A}^i} = \frac{\partial \mathcal{L}}{\partial (\mathbf{B}_A)^i} \cdot \frac{\partial (\mathbf{B}_A)^i}{\partial (\mathbf{A})^i} \approx \frac{\partial \mathcal{L}}{\partial (\mathbf{B}_A)^i} \cdot \frac{\partial F(\mathbf{A}^i)}{\partial \mathbf{A}^i},
\end{equation}
where $\dfrac{\partial F(\mathbf{A}^i)}{\partial \mathbf{A}^i}$ is defined by
\begin{equation}
    \dfrac{\partial F(\mathbf{A}^i)}{\partial \mathbf{A}^i}= \left\{
    \begin{array}{ll}
        2 + 2\mathbf{A}^i, &\text{ if } -1 \le \mathbf{A}^i < 0, \\
        2 - 2\mathbf{A}^i, &\text{ if }\quad\, 0  \le \mathbf{A}^i < 1, \\
        0, &\text{ otherwise. } 
    \end{array}\right.
\end{equation}

Besides, the STE~\cite{bengio2013estimating} is used to calculate the gradient of the loss $\mathcal{L}$ $\emph{w.r.t.}$ the weight $\mathbf{w}_j^i$ as
\begin{equation}\label{gradient_weight}
  \frac{\partial \mathcal{L}}{\partial \mathbf{w}_j^i} = \frac{\partial \mathcal{L}}{\partial (\mathbf{b}_w)_j^i} \cdot \frac{\partial (\mathbf{b}_w)_j^i}{\partial \mathbf{w}_j^i} \approx \frac{\partial \mathcal{L}}{\partial (\mathbf{b}_w)_j^i}.
\end{equation}

\textbf{BNN Inference}. In practical deployment, the BNN model is accelerated using the efficient XNOR and bitcount logics embeded in the hardware. Thus, the quantized weights and activations need to be further transformed back into $\{ 0, 1 \}$ space. Such a transformation process can be realized by setting
\begin{align}
  & (\bar{\mathbf{B}}_A)^i = \big( 1 + (\mathbf{B}_A)^i \big) / 2.\label{activation_01}\\
  & (\bar{\mathbf{b}}_w)^i_j = \big(1 + (\mathbf{b}_w)^i_j \big) / 2.\label{weight_01}
\end{align}

Then, the approximated convolution in Eq.\,(\ref{approximated_convolution}) can be replaced by the following equality
\begin{equation}\label{equal}
\begin{split}
 \mathbf{a}^{i+1}_j \approx {\beta}^i_j \cdot \big( 2 \cdot (\bar{\mathbf{b}}_w)_j^i \odot (\bar{\mathbf{B}}_A)^i - n^i\big),
\end{split}
\end{equation}
where $\odot$ represents the XNOR and bitcount operations that are well-fitted for real-time network inference.

\textbf{Our Insight}. In this paper, we focus on binarizing the real-valued weight $\mathbf{w}_j^i$. Different from most existing works~\cite{hubara2016binarized,zhou2016dorefa,bulat2019xnor,lin2020rotated} that project weights $\mathbf{w}_j^i$ into $(\mathbf{b}_w)_j^i \in \{-1, +1\}^{n^i}$ using the sign function during training and then transform $(\mathbf{b}_w)_j^i$ into $(\bar{\mathbf{b}}_w)_j^i \in \{0, +1\}^{n^i}$ for inference, we seek to directly encode the weights into $\bar{\mathbf{b}}_w \in \{ 0, +1 \}^{n^i}$ and then devise an efficient optimization to attain the optimal solution in Sec.\,\ref{learning_objective}. We demonstrate in Sec.\,\ref{weight_distribution} that the weight, $\mathbf{w}_j^i$, roughly follows a Laplacian distribution, which inhibits the entropy maximization. We reveal that this can be easily addressed by removing the $\ell_2$ regularization in Sec.\,\ref{maximizing_bit_entropy}.

For simplicity, the scripts ``$i$'' and ``$j$'' are omitted in the following context.

\section{Weight Binarization}
\label{approach}

In this section, we specify the formulation of our weight binarization, including the binary learning objective, weight distribution, and bit entropy maximization.

\subsection{Learning Objective\label{learning_objective}}

To achieve high-quality weight binarization, different from the conference version~\cite{lin2020rotated}, we reformulate the learning objective in Eq.\,(\ref{rotation}) as
\begin{equation}\label{objective}
\begin{split}
 &\mathop{\arg\max}_{\bar{\mathbf{b}}_w} \frac{(\bar{\mathbf{b}}_w)^T |\mathbf{w}|}{\| \bar{\mathbf{b}}_w \|_2 \big\| |\mathbf{w}| \big\|_2}, 
  \\&
\;\; s.t. \quad \bar{\mathbf{b}}_w \in \{0, +1\}^n,
\end{split}
\end{equation}
where $| \cdot |$ returns the absolute result of its input.

As can be seen, our learning objective is also built on the basis of angle alignment. Nevertheless, our method differs from Eq.\,(\ref{rotation}) in many aspects: First, we drop the sign function since variables in a binarized network must be retained in a discrete set; thus, the binarization should be built upon the concept of the discrete optimization rather than the simple sign function. Second, we encode the weights into $\bar{\mathbf{b}}_w \in \{0, +1\}^n$ rather than $\mathbf{b}_w \in \{-1, +1\}^n$. In Corollary 1, we show that $\bar{\mathbf{b}}_w \in \{0, +1\}^n$ allows us to find an analytical solution in an efficient manner by transferring the high-magnitude weights to $+1$s and $0$s otherwise. Third, our angle alignment is independent of the rotation matrix, $\mathbf{R}$, since we remove the sign function, which makes the rotation direction unpredictable. Lastly, we align the angle difference between the binarization and absolute weight vector, $|\mathbf{w}|$, instead of the weight $\mathbf{w}$ itself. The rationale behind this is that our binarization falls into the non-negative set $\bar{\mathbf{b}}_w$. Fig.\,\ref{bnn}(b) outlines our binarization process.

Note that $\big\| |\mathbf{w}| \big\|_2$ is irrelevant to the optimization of Eq.\,(\ref{objective}). Thus, the learning can be simplified to
\begin{equation}\label{bw}
\begin{split}
   &\mathop{\arg\max}_{\bar{\mathbf{b}}_w} \; \frac{(\bar{\mathbf{b}}_w)^T}{\| \bar{\mathbf{b}}_w \|_2} |\mathbf{w}|, 
  \\&
    s.t. \quad \bar{\mathbf{b}}_w \in \{0, +1\}^n.
\end{split}
\end{equation}

This is an integer programming problem~\cite{conforti2014integer}. Nevertheless, as demonstrated in Corollary 1, by learning the encoding space in $\{0, +1\}^n$, we can reach the global maximum in a substantially efficient fashion.

\textbf{Corollary 1.} For Eq.\,(\ref{bw}), the computational complexity of finding the global optimum is $\mathcal{O}(n\log{n})$.

\textbf{Proof:} For $\bar{\mathbf{b}}_w \in \{ 0, +1 \}^n$, it is intuitive to see that $\| \bar{\mathbf{b}}_w \|_2$ falls into the set $\{ \sqrt{1}, ..., \sqrt{n} \}$. Considering that $\| \bar{\mathbf{b}}_w \|_2 = \sqrt{k}$ ($k = 1, ..., n$), the integer programming problem in Eq.\,(\ref{bw}) can be maximized by encoding to $+1$s those elements of $\bar{\mathbf{b}}_w$ that correspond to the largest $k$ entries of $|\mathbf{w}|$. To this end, we need to perform sorting upon $|\mathbf{w}|$, for which the complexity is $\mathcal{O}(n\log{n})$. Since $k$ has $n$ possible values, we need to evaluate Eq.\,(\ref{bw}) $n$ times, and then select the $\bar{\mathbf{b}}_w$ that maximizes the objective function, leading to a linear complexity  with $n$. Hence, the overall complexity is $\mathcal{O}(n\log{n})$. $\hfill\blacksquare$

Therefore, given one filter weight $\mathbf{w} \in \mathbb{R}^n$, we can find the binarization $\bar{\mathbf{b}}_w \in \{0, +1\}^n$ having the smallest angle with $|\mathbf{w}|$. Note that $\bar{\mathbf{b}}_w$ found in this way is the global optimum. Furthermore, we emphasize that the proof of Corollary 1 indicates that the binarization in our framework involves the magnitudes of weights instead of the signs of weights, which significantly differentiates our work from existing works. In the next two sections, we show that the overall complexity can be further reduced to $\mathcal{O}(n)$, given the bit entropy maximization.

\subsection{Weight Distribution\label{weight_distribution}}

The capacity of a binarization model, often measured by the bit entropy, is maximized when it is half-half distributed, \emph{i.e.}, one half of the weights are encoded into $0$ and the other half are encoded into $+1$~\cite{lin2020rotated,qin2020forward}. In this case, we expect to maximize our objective in Eq.\,(\ref{bw}) when those weights with the top-half magnitudes are encoded into $+1$s and the remaining are encoded into $0$s. However, we reveal that it is difficult to binarize $\mathbf{w}$ with entropy maximization due to its specific form of distribution.

Specifically, after training, $w \in \mathbf{w}$ is widely believed to roughly obey a zero-mean Laplacian distribution, \emph{i.e.}, $w \sim La(0, b)$, or a zero-mean Gaussian distribution, \emph{i.e.}, $w \sim \mathcal{N}(0, \sigma^2)$~\cite{cai2017deep,banner2018post,zhong2020towards}. In Corollary 2, for the first time, we demonstrate its specific distribution.

\begin{figure*}[!t]
  \centering
  \begin{subfigure}{0.46\linewidth}
    \includegraphics[width=\linewidth]{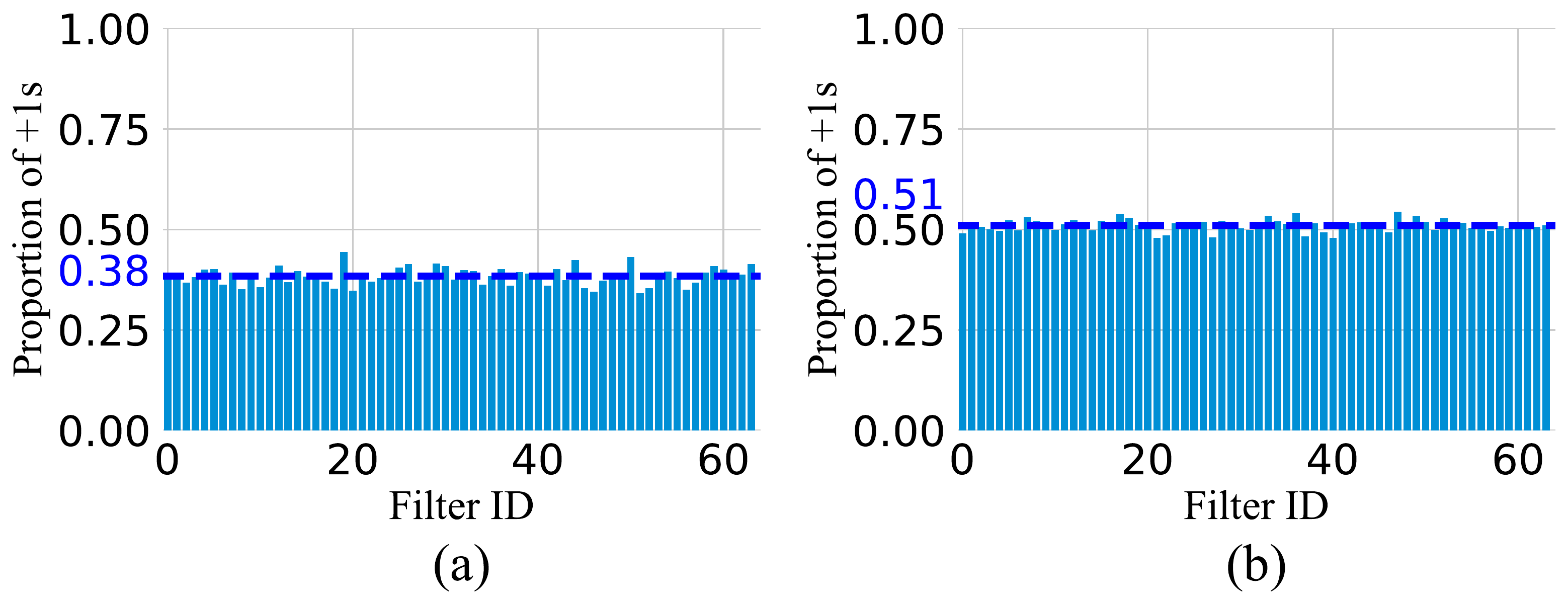}
    \caption{Layer1.1.2 of ResNet-18}
    \label{phenomenon:real}
  \end{subfigure}
  \hspace{.3in}
  \vspace{.15in}
  \begin{subfigure}{0.46\linewidth}
    \includegraphics[width=\linewidth]{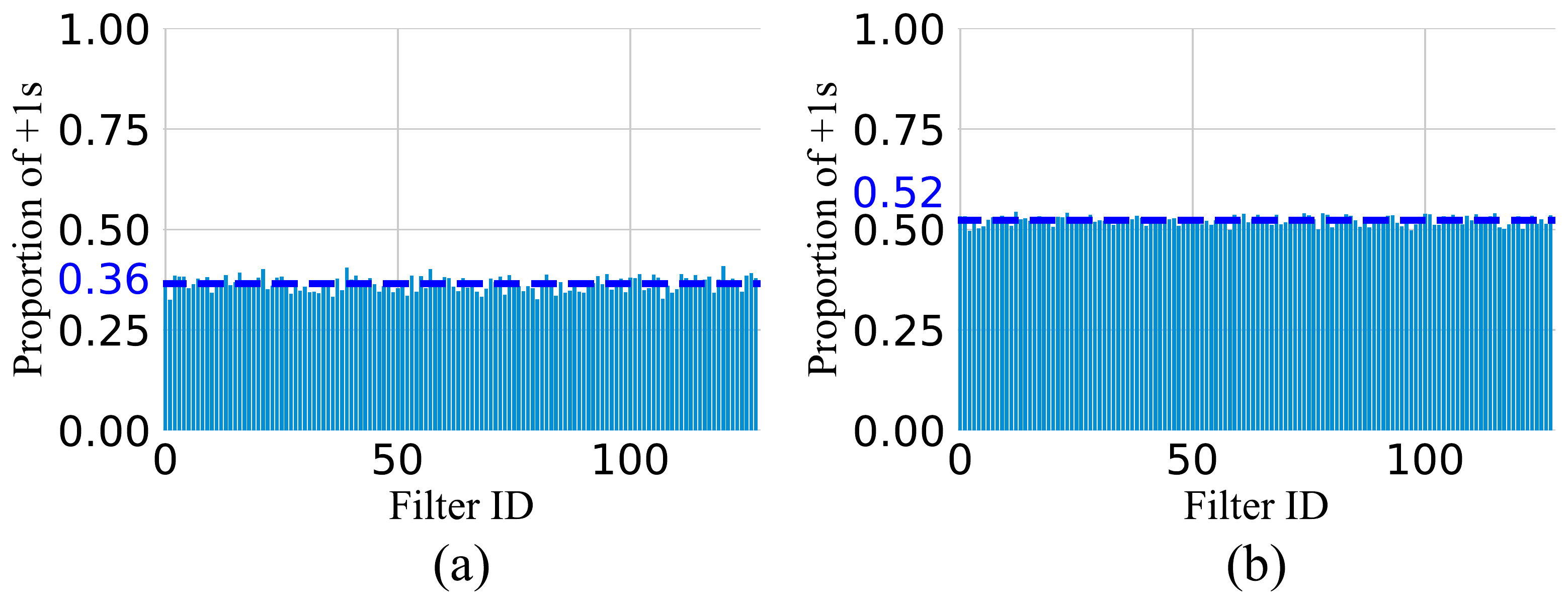}
    \caption{Layer2.1.2 of ResNet-18}
    \label{phenomenon:zero}
  \end{subfigure} \\
  \begin{subfigure}{0.46\linewidth}
    \includegraphics[width=\linewidth]{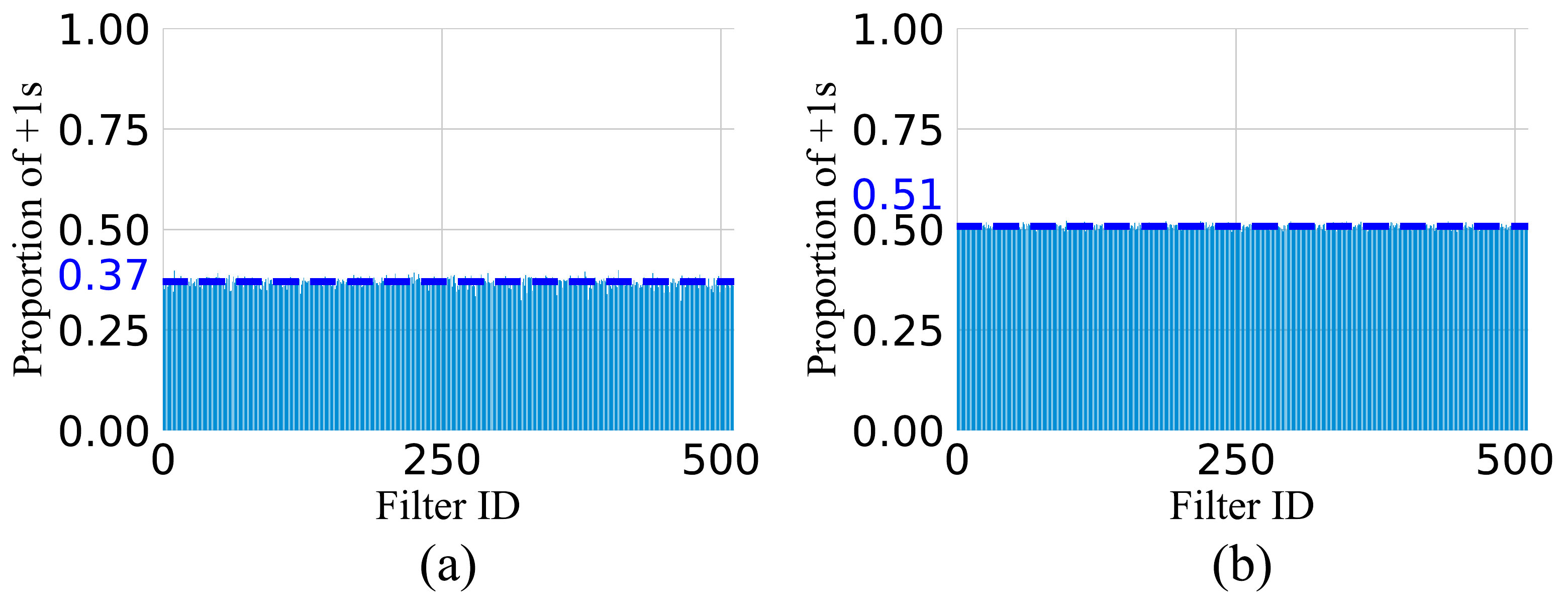}
    \caption{Layer4.1.2 of ResNet-18}
    \label{phenomenon:dsg}
  \end{subfigure}
  \hspace{.3in}
  \vspace{.15in}
  \begin{subfigure}{0.46\linewidth}
    \includegraphics[width=\linewidth]{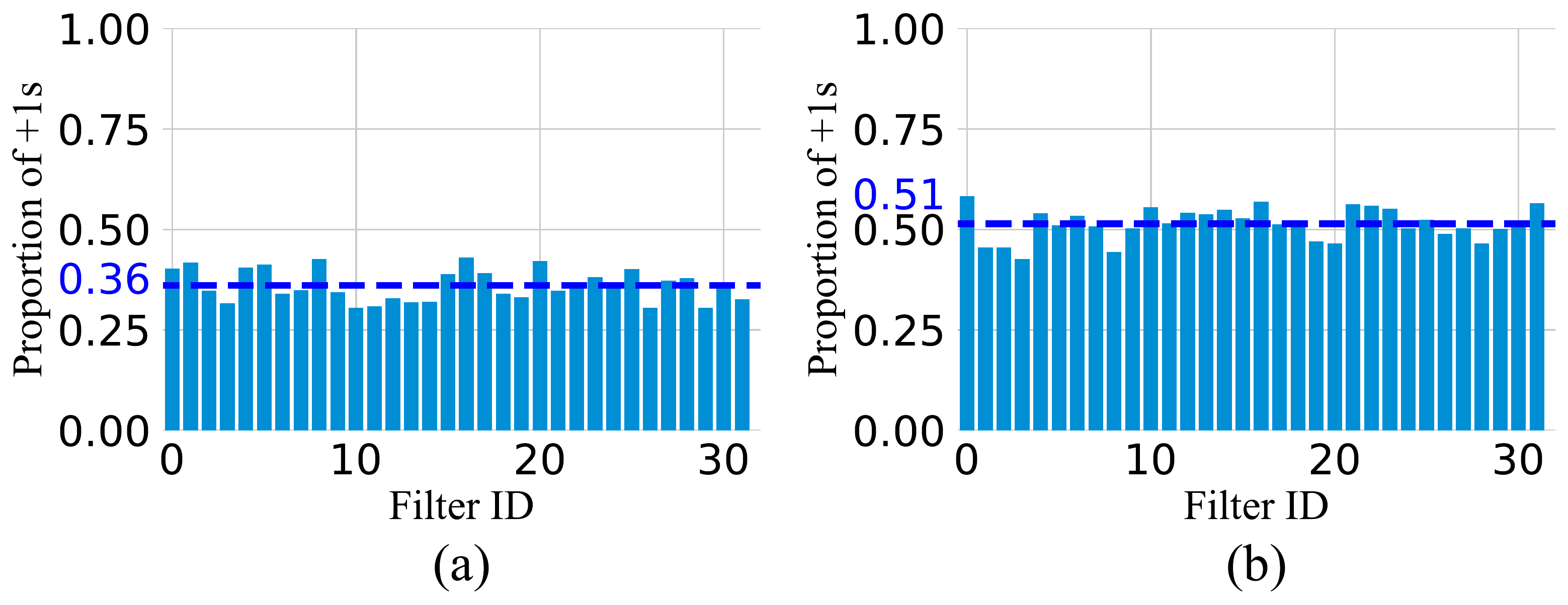}
    \caption{Layer2.1.2 of ResNet-20} 
    \label{phenomenon:zeroq+IL}
  \end{subfigure}  \\
  \begin{subfigure}{0.46\linewidth}
    \includegraphics[width=\linewidth]{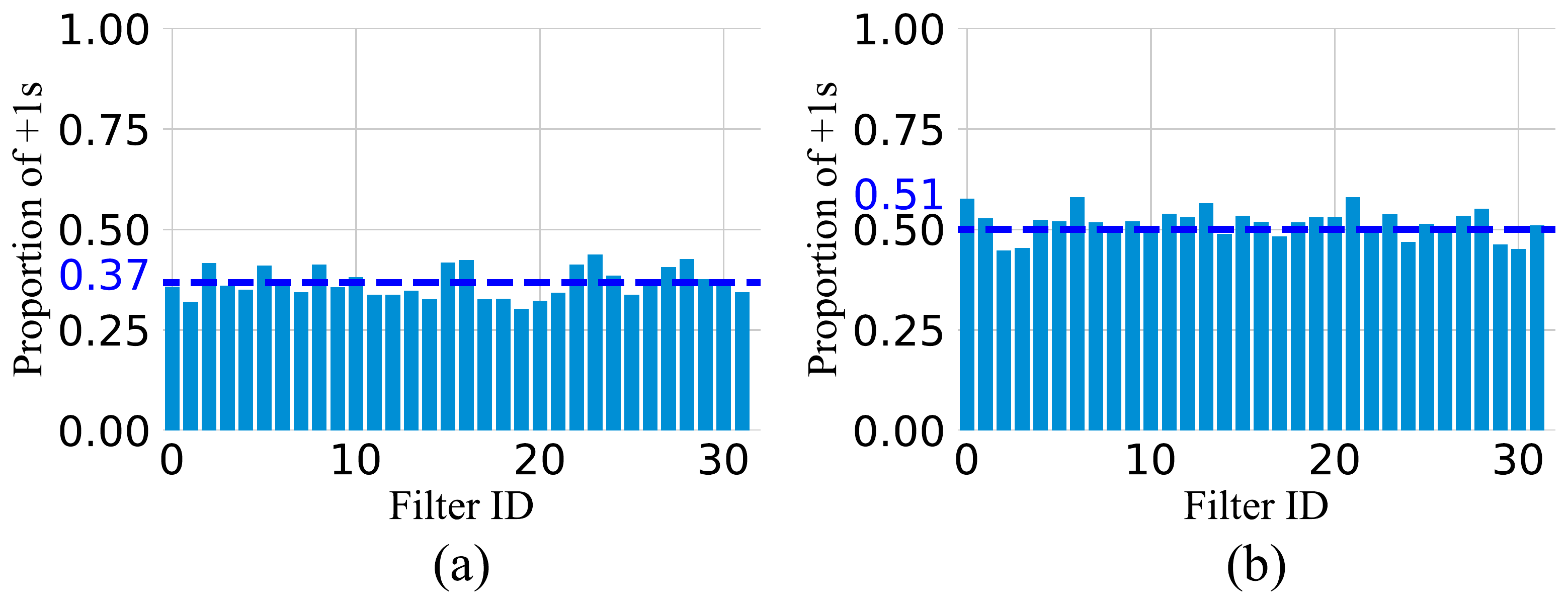}
    \caption{Layer2.2.2 of ResNet-20}
    \label{phenomenon:dsg+IL}
  \end{subfigure}
  \hspace{.3in}
  \begin{subfigure}{0.46\linewidth}
    \includegraphics[width=\linewidth]{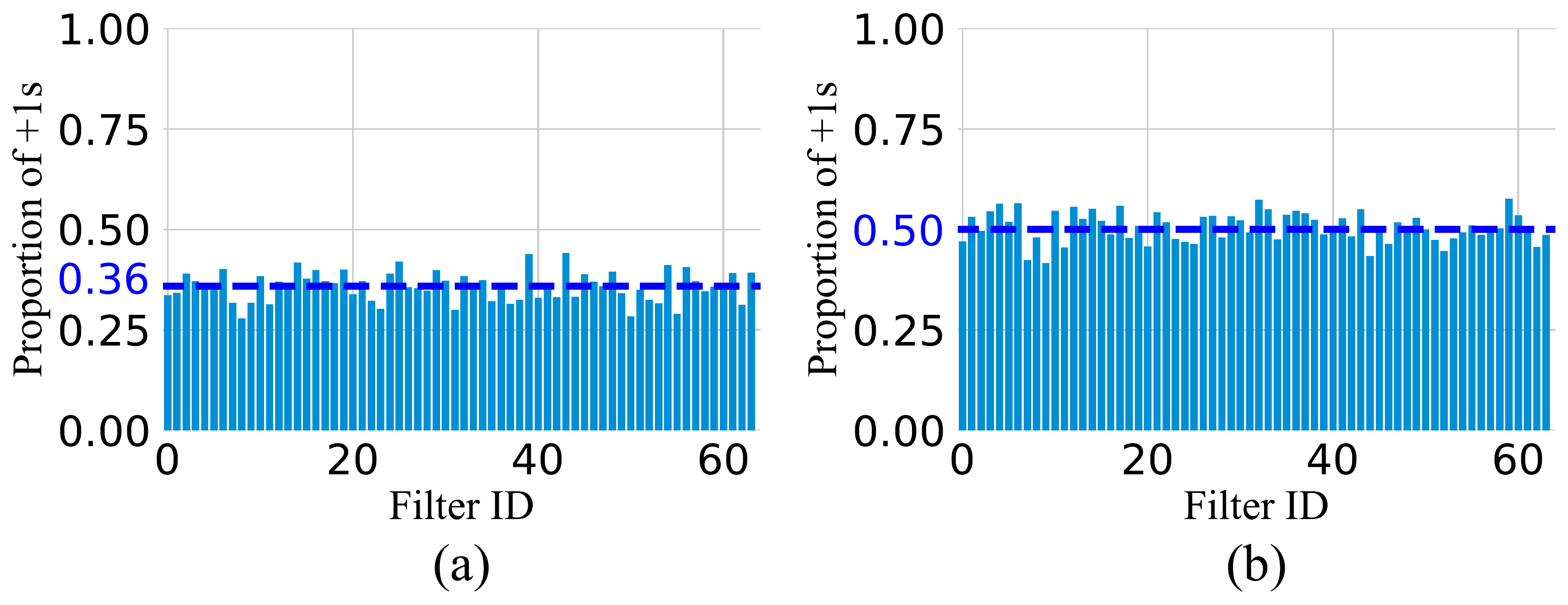}
    \caption{Layer3.2.1 of ResNet-20}
    \label{phenomenon:ours}
  \end{subfigure}
  \caption{Proportion of $+1$s trained (a) with and (b) without the $\ell_2$ regularization across different layers in different networks. The dashed blue lines denote the average proportions of $+1$s of all filter weights.}
  \vspace{-1.0em}
  \label{ratio}
\end{figure*}

\textbf{Corollary 2.} $w \in \mathbf{w}$ roughly follows a zero-mean Laplacian distribution.

\textbf{Proof:} Suppose $w$ is encoded into $+1$ if $|w| > t$, and 0 otherwise. Note that the learning of Eq.\,(\ref{bw}) can actually be regarded as a problem of finding the centroid of a subset~\cite{shakarji2013theory} as well. Consequently, the learning process can be calculated by the integral as
\begin{equation}\label{centroid}
\begin{split}
 \frac{(\bar{\mathbf{b}}_w)^T}{\| \bar{\mathbf{b}}_w \|_2} |\mathbf{w}| &= \frac{\int^{-t}_{-\infty}wf(w)dw + \int_{t}^{+\infty}wf(w)dw}{\sqrt{\int^{-t}_{-\infty}f(w)dw + \int_{t}^{+\infty}f(w)dw}}
 \\&
 = \frac{2\int_t^{+\infty}wf(w)dw}{\sqrt{2\int_t^{+\infty}f(w)dw}},
\end{split}
\end{equation}
where $f(w)$ represents the probability density function with regard to $w$. {Then, we denote} $p_{+1}$ as the proportion of $\mathbf{w}$ being encoded into $+1$s. Intuitively, its calculation can be derived as
\begin{equation}
 p_{+1} = 1 - 2\int^t_0 f(w)dw.
\end{equation}

To demonstrate our corollary, we first derive the theoretical values of $p_{+1}$ when $p(w)$ follows a Laplacian or Gaussian distribution, and then experimentally complete our final proof.

\textit{Laplacian distribution}. In this case, we have $f(w) = \frac{1}{2b}e^{-w/b}$. Therefore,  Eq.\,(\ref{centroid}) becomes
\begin{equation}\label{laplace}
\begin{split}
 \frac{(\bar{\mathbf{b}}_w)^T}{\| \bar{\mathbf{b}}_w \|_2} |\mathbf{w}| &
 = \frac{2\int_t^{+\infty}\frac{w}{2b}e^{-w/b}dw}{\sqrt{2\int_t^{+\infty}\frac{1}{2b}e^{-w/b}dw}}
 \\&
 = \frac{(b + t)e^{-t/b}}{\sqrt{e^{-t/b}}} 
 \\&
 = (b + t)\sqrt{e^{-t/b}}.
\end{split}
\end{equation}

Setting $\frac{\partial (b + t)\sqrt{e^{-t/b}}}{\partial t} = 0$ to attain the maximum of Eq.\,(\ref{laplace}), we have $t = b$. The proportion of $+1$s can be obtained as
\begin{equation}
p_{+1} = 1 - 2\int_0^t\frac{1}{2b}e^{-w/b}dw \approx 0.37.
\end{equation}

\textit{Gaussian distribution}. In this case, we have $f(w) = \frac{1}{\sqrt{2\pi}\sigma}e^{-w^2/(2\sigma^2)}$. Similarly, Eq.\,(\ref{centroid}) can be rewritten as
\begin{equation}\label{gaussian}
\begin{split}
 \frac{(\bar{\mathbf{b}}_w)^T}{\| \bar{\mathbf{b}}_w \|_2} |\mathbf{w}| &= \frac{2\int_t^{+\infty}\frac{w}{\sqrt{2\pi}\sigma}e^{-w^2/(2\sigma^2)}dw}{\sqrt{2\int_t^{+\infty}\frac{1}{\sqrt{2\pi}\sigma}e^{-w^2/(2\sigma^2)}dw}}
 \\&
 = \frac{\frac{\sigma}{\sqrt{2\pi}}e^{-t^2/(2\sigma^2)}}{\sqrt{\frac{1}{2}\operatorname{erfc}(\frac{t}{\sqrt{2}\sigma})}},
\end{split}
\end{equation}
where $\operatorname{erfc}(\cdot)$ represents the well-known complementary error function~\cite{andrews1998special}.

Let $m = \frac{t}{\sqrt{2}\sigma}$, then Eq.\,(\ref{gaussian}) can be written as
\begin{equation}
 \frac{(\bar{\mathbf{b}}_w)^T}{\| \bar{\mathbf{b}}_w \|_2} |\mathbf{w}| = \frac{\sigma}{\sqrt{\pi}}\frac{e^{-m^2}}{\sqrt{\operatorname{erfc}(m)}}.
\end{equation}

Setting $\frac{\partial e^{-m^2}/\sqrt{\operatorname{erfc}(m)}}{\partial m} = 0$, we have $m = \frac{t}{\sqrt{2}\sigma} \approx 0.43$. Thus, we obtain $t \approx 0.43\sqrt{2}\sigma$, and then derive the proportion of $+1$s:
\begin{equation}
 p_{+1} = 1 - 2 \int_0^t\frac{1}{\sqrt{2\pi}\sigma}e^{-w^2/(2\sigma^2)}dw \approx 0.54.
\end{equation}

As mentioned above, the trained weight $w \in \mathbf{w}$  obeys either a Laplacian distribution (with $p_{+1} \approx 0.37$) or a Gaussian distribution (with $p_{+1} \approx 0.54$). In Fig.\,\ref{ratio}(a), we conduct an experiment which shows a practical $p_{+1}$ of around 0.36$\sim$0.38 after training.\footnote{Similar phenomena can be observed in other layers and networks as well.} This implies that $w \in \mathbf{w}$ follows a Laplacian distribution, and then our proof is completed. $\hfill\blacksquare$

\subsection{Maximizing Bit Entropy \label{maximizing_bit_entropy}}

The proof of Corollary 1 indicates that the binarization in our framework is related to the weight magnitude, \emph{i.e.}, $|\mathbf{w}|$. However, the Laplacian distribution contradicts the entropy maximization. To solve this, one naive solution is to assign the top half of elements of the sorted $|\mathbf{w}|$ with $+1$  and assign the remaining elements with $0$, that is
\begin{equation}\label{half-half}
    \tilde{\mathbf{b}}_w =\left\{\begin{array}{l}
        +1, \text{ top half of sorted } |\mathbf{w}|, \\
        0, \text{ otherwise.}
    \end{array}\right. 
\end{equation}

%

Despite helping achieve entropy maximization, such a simple operation violates the learning objective of minimizing the angular bias in Eq.\,(\ref{bw}) since $\tilde{\mathbf{b}}_w$ deviates significantly from the optimal $\bar{\mathbf{b}}_w$ as revealed in Corollary 3.

\textbf{Corollary 3.} Suppose $\bar{\mathbf{b}}_w$ is a binarized vector with a total of $k$ $+1$s and the binarized vector $\tilde{\mathbf{b}}_w$ has $r$ different bits from $\bar{\mathbf{b}}_w$. Then the angle between $\bar{\mathbf{b}}_w$ and $\tilde{\mathbf{b}}_w$ is bounded by $\big[ \arccos\sqrt{\frac{k}{k+r}}, \arccos\sqrt{\frac{k-r}{k}} \big]$.

\textbf{Proof:} To find the lower bound, we need to obtain $\tilde{\mathbf{b}}_w$ such that $\frac{(\bar{\mathbf{b}}_w)^T\tilde{\mathbf{b}}_w}{\| \bar{\mathbf{b}}_w \|_2 \| \tilde{\mathbf{b}}_w \|_2}$ is maximized. Intuitively, this can be achieved when $\tilde{\mathbf{b}}_w$ has all ones at the same positions as $\bar{\mathbf{b}}_w$, and $r$ additional ones in the remaining positions, in which $\| \tilde{\mathbf{b}}_w \|_2 = \sqrt{k + r}$ and $(\bar{\mathbf{b}}_w)^T\tilde{\mathbf{b}}_w = k$. Then, we have the lower bound of $\arccos \sqrt{\frac{k}{k+r}}$. To obtain the upper bound, we need to minimize $\frac{(\bar{\mathbf{b}}_w)^T\tilde{\mathbf{b}}_w}{\| \bar{\mathbf{b}}_w \|_2 \| \tilde{\mathbf{b}}_w \|_2}$, which can be done when there are ($k - r$) $+1$s in $\tilde{\mathbf{b}}_w$ in the common positions with $\bar{\mathbf{b}}_w$ and the rest are set to zeros. In this case, we have $\| \tilde{\mathbf{b}} \|_2 = \sqrt{k - r}$ and $(\bar{\mathbf{b}}_w)^T\tilde{\mathbf{b}}_w = k - r$, which leads to the upper bound of $\arccos \sqrt{\frac{k - r}{k}}$.

According to Corollary 2 and Eq.\,(\ref{half-half}), we have $k \approx 0.37n$, and $r \approx 0.13n$. Then, we can derive the practical angle bounds as $[\arccos \sqrt{\frac{0.37}{0.37 + 0.13}}, \arccos \sqrt{\frac{0.37 - 0.13}{0.37}}] \approx [30.66^{\circ}, 36.35^{\circ}]$. Therefore, a large angle bias occurs between the solution $\tilde{\mathbf{b}}_w$ from Eq.\,(\ref{half-half}) and the solution $\bar{\mathbf{b}}_w$ from optimizing Eq.\,(\ref{bw}). In Sec.\,\ref{ablation_studies}, we demonstrate the poor performance when simply using $\tilde{\mathbf{b}}_w$.

\begin{algorithm}[!t]
   \caption{Sign-to-Magnitude Network Binarization}
   \label{alg1}
\begin{algorithmic}
   \STATE {\bfseries Input:} An $L$-layer full-precision network with weights $\mathbf{W}^i = \{\mathbf{w}_1^i, \mathbf{w}_2^i, ..., \mathbf{w}^i_{c^i_{out}} \}$ ($i = 1, 2,..., L$), input images (activations) $\mathbf{A}^1 = \{\mathbf{a}_1^1, \mathbf{a}_2^1, ..., \mathbf{a}^1_{c_{in}^1} \}$.
   \STATE \textbf{1) Forward Propagation:}
   \STATE Remove the $\ell_2$ regularization term.
   \FOR{$i = 1$ {\bfseries to} $L$}
       \STATE Binarize the inputs $(\mathbf{B}_A)^i = \operatorname{sign}(\mathbf{A}^i)$ (Eq.\,(\ref{binarization}));
        \FOR{$j = 1$ {\bfseries to} $c^i_{out}$}
         \STATE Obtain the half-half binarization $(\tilde{\mathbf{b}}_w)^i_j$ (Eq.\,(\ref{half-half}));
         \STATE Obtain the binarization $(\mathbf{b}_w)^i_j = 2\cdot(\tilde{\mathbf{b}}_w)^i_j - 1$ (the inverse of Eq.\,(\ref{weight_01}));
         \STATE Conduct the convolution $\mathbf{a}^{i+1}_j \approx {\beta}^i_j \cdot (\mathbf{b}_w)_j^i \circledast (\mathbf{B}_A)^i$ (Eq.\,(\ref{approximated_convolution}));
        \ENDFOR
        \STATE $\mathbf{A}^{i+1} = \{\mathbf{a}_1^{i+1}, \mathbf{a}_2^{i+1}, ..., \mathbf{a}_{c_{out}^{i+1}}^{i+1} \}$;
   \ENDFOR
   \STATE \textbf{2) Backward Propagation:}
   \FOR{$i = L$ {\bfseries to} $1$}
       \STATE Compute gradient $\frac{\partial \mathcal{L}}{\partial \mathbf{A}^i} \approx \frac{\partial \mathcal{L}}{\partial (\mathbf{B}_A)^i} \cdot \frac{\partial F(\mathbf{A}^i)}{\partial \mathbf{A}^i}$ (Eq.\,(\ref{gradient_activation}));
       \FOR{$j = 1$ {\bfseries to} $c^i_{out}$}
        \STATE Compute gradient $\frac{\partial\mathcal{L}}{\partial \mathbf{w}_j^i} \approx \frac{\partial \mathcal{L}}{\partial (\mathbf{b}_w)^i_j}$ (Eq.\,(\ref{gradient_weight}));
       \ENDFOR       
   \ENDFOR
   \STATE \textbf{3) Weight Updating:}
   \FOR{$i = L$ {\bfseries to} $1$}
       \FOR{$j = 1$ {\bfseries to} $c^i_{out}$}
        \STATE Update $\mathbf{w}^i_j = \mathbf{w}^i_j - \eta \frac{\partial\mathcal{L}}{\partial \mathbf{w}_j^i}$; \# $\eta$ denotes learning rate
       \ENDFOR       
   \ENDFOR
   \STATE {\bfseries Output:} An $L$-layer binarized network with weights $(\tilde{\mathbf{b}}_W)^i = \{(\tilde{\mathbf{b}}_w)^i_1, (\tilde{\mathbf{b}}_w)^i_2, ..., (\tilde{\mathbf{b}}_w)^i_{c^i_{out}}\} (i = 1, 2, ..., L)$.
\end{algorithmic}
\end{algorithm}

Instead of imposing an additional loss term to regularize the ideal half-half binarization, we analyze the numerical value of each weight and reveal that simply removing the $\ell_2$ regularization can explicitly maximize the bit capacity, leading to a more informative binarized network.

Let $\mathcal{L}^k_{\mathbf{b}_w} =\mathop{\arg\max}_{\mathbf{b}_w} \frac{(\mathbf{b}_w)^T}{\sqrt{k}}|\mathbf{w}| = \frac{\sum_{i=1}^k\tilde{w}_i}{\sqrt{k}}$ denote the maximum result of the integer programming problem in Eq.\,(\ref{bw}), where $\tilde{w}_i \in |\mathbf{w}|$ corresponds to the $i$-th largest magnitude. We have $\mathcal{L}^{k+1}_{\mathbf{b}_w} < \mathcal{L}^{k}_{\mathbf{b}_w}$, \emph{i.e.},
\begin{equation}
   \frac{ \tilde{w}_{k+1} + \sum_{i=1}^{k} \tilde{w}_i}{ (\sqrt{k+1} - \sqrt{k}) + \sqrt{k}} < \frac{ \sum_{i=1}^{k} \tilde{w}_i}{\sqrt{k}}.
\end{equation}

We can deduce that
%
\begin{equation}\label{inequality}
    \tilde{w}_{k+1} < \mathcal{L}^k_{\mathbf{b}_w}(\sqrt{k+1} - \sqrt{k}).
\end{equation}

For Laplacian distributed weights, we know that $k \approx 0.37n$. Thus, the above inequality can be rewritten as
\begin{equation}\label{inequality2}
    \tilde{w}_{k+1} < \mathcal{L}^k_{\mathbf{b}_w}(\sqrt{0.37n+1} - \sqrt{0.37n}).
\end{equation}

Since $n$ is typically thousands for a neural network and we statistically find that  $\mathcal{L}^k_{\mathbf{b}_w}$ ranges from 0.63 to 0.73, the multiplication of the two terms in Eq.\,(\ref{inequality2}) thus results in an extremely small $\tilde{w}_{k+1}$ approximating to zero. Thus, we need to enlarge the value of $\tilde{w}_{k + 1}$ to break the above inequality. We realize that one of the major causes for a small $\tilde{w}_{k+1}$ lies in the existence of the $\ell_2$ regularization imposed on the training of neural network. This inspires us to remove the $\ell_2$ regularization for the to-be-binarized weights $\mathbf{w}$. 

As shown in Fig.\,\ref{ratio}(b), the removal of the $\ell_2$ regularization increases the proportion of $+1$s in $\bar{\mathbf{b}}_w$ to around 0.50$\sim$0.52. Taking 0.51 as an example, we then further enforce the ideal half-half binarization $\tilde{\mathbf{b}}_w$ using Eq.\,(\ref{half-half}). As a result, $k \approx 0.51n$ and $r \approx 0.01n$, yielding the much smaller angle bounds of $[\arccos \sqrt{\frac{0.51}{0.51 + 0.01}}, \arccos \sqrt{\frac{0.51 - 0.01}{0.51}}] \approx [7.97^{\circ}, 8.05^{\circ}]$ between $\tilde{\mathbf{b}}_w$ and $\bar{\mathbf{b}}_w$. This effectively increases the bit entropy and leads to a nearly optimal solution for the learning objective of Eq.\,(\ref{bw}). Besides, the half-half binarization further reduces the computational complexity of $\mathcal{O}(n \log n)$ to $\mathcal{O}(n)$ since we only need to find the median of $|\mathbf{w}|$, and encode weights into $+1$s when their magnitude is larger than the median, and 0s otherwise.

The forward and backward processes of SiMaN are summarized in Algorithm\,\ref{alg1}. During training, we remove the $\ell_2$ regularization  and adopt the binarization $\mathbf{b}_w \in \{-1, +1\}^n$ transformed from the half-half binarization $\tilde{\mathbf{b}}_w \in \{0, +1\}^n$ for the convolution in Eq.\,(\ref{approximated_convolution}). After training, we obtain a network consisting of a binarized weight $\tilde{\mathbf{b}}_w$ for practical deployment on hardware where the convolution is executed using the XNOR and bitcount operations in Eq.\,(\ref{equal}).

\section{Experiments}

To demonstrate the efficacy of the proposed SiMaN binarization scheme, we compare its performance with several state-of-the-art BNNs~\cite{courbariaux2016binarized,zhou2016dorefa,rastegari2016xnor,lin2017towards,cai2017deep,wan2018tbn,liu2018bi,ding2019regularizing,gong2019differentiable,gu2019bayesian,gu2019projection,yang2020searching,wang2020sparsity,qin2020forward,han2020training} as well as the conference version~\cite{lin2020rotated}
on two image classification datasets, including CIFAR-10~\cite{krizhevsky2009learning} and ImageNet~\cite{deng2009imagenet}.

\subsection{Datasets and Experimental Settings}

\textbf{CIFAR-10}~\cite{krizhevsky2009learning} consists of $60,000$ $32\times32$ images from $10$ classes. Each class has $6,000$ images. We split the dataset into $50,000$ training images and $10,000$ testing images. Data augmentation includes random cropping and random flipping, as done in~\cite{he2016deep} for the training images.

\textbf{ImageNet}~\cite{deng2009imagenet} contains over $1.2$ million images for training and $50,000$ validation images from $1,000$ classes for classification. For fair comparison with the recent advances in~\cite{qin2020forward,han2020training,lin2020rotated}, we only apply the data augmentation including random cropping and flipping.

\textbf{Network Structures} For CIFAR-10, we binarize ResNet-18/20~\cite{he2016deep} and VGG-small~\cite{zhang2018lq}. For ImageNet, ResNet-18/34 are chosen for binarization. Following~\cite{qin2020forward,han2020training,lin2020rotated}, double skip connections~\cite{liu2018bi} are added to the ResNets and we do not binarize the first and last layers for all networks.

\textbf{Implementation Details} We implement our SiMaN using Pytorch~\cite{paszke2019pytorch} and all experiments are conducted on NVIDIA Tesla V100 GPUs. We use the cosine scheduler with a learning rate of $0.1$~\cite{qin2020forward,lin2020rotated}. The SGD is adopted as the optimizer with a momentum of $0.9$. For those layers that are not binarized, the weight decay is set to $5\times 10^{-4}$ on CIFAR-10 and $1 \times 10^{-4}$ on ImageNet, and $0$ otherwise to remove the $\ell_2$ regularization for the bit entropy maximization as discussed in Sec.\,\ref{maximizing_bit_entropy}. We train the models from scratch with $400$ epochs and a batch size of $256$ on CIFAR-10, and with $150$ epochs and a batch size of $512$ on ImageNet.

Note that, we only apply the classification loss during network training for fair comparison. Other training losses such as those proposed in~\cite{hou2016loss,ding2019regularizing,wang2019learning}, the variants of network structures in~\cite{bethge2020meliusnet,zhu2019binary,liu2020reactnet}, and even the two-step training strategy~\cite{martinez2019training} can be integrated to further boost the binarized networks' performance. These, however, are not considered here. We aim to show the advantages of our magnitude-based optimization solution over the traditional sign-based methods under regular training loss, the same network structure and a common training strategy.

\begin{figure*}[!t]
  \centering
  \begin{subfigure}{0.43\linewidth}
    \includegraphics[width=\linewidth]{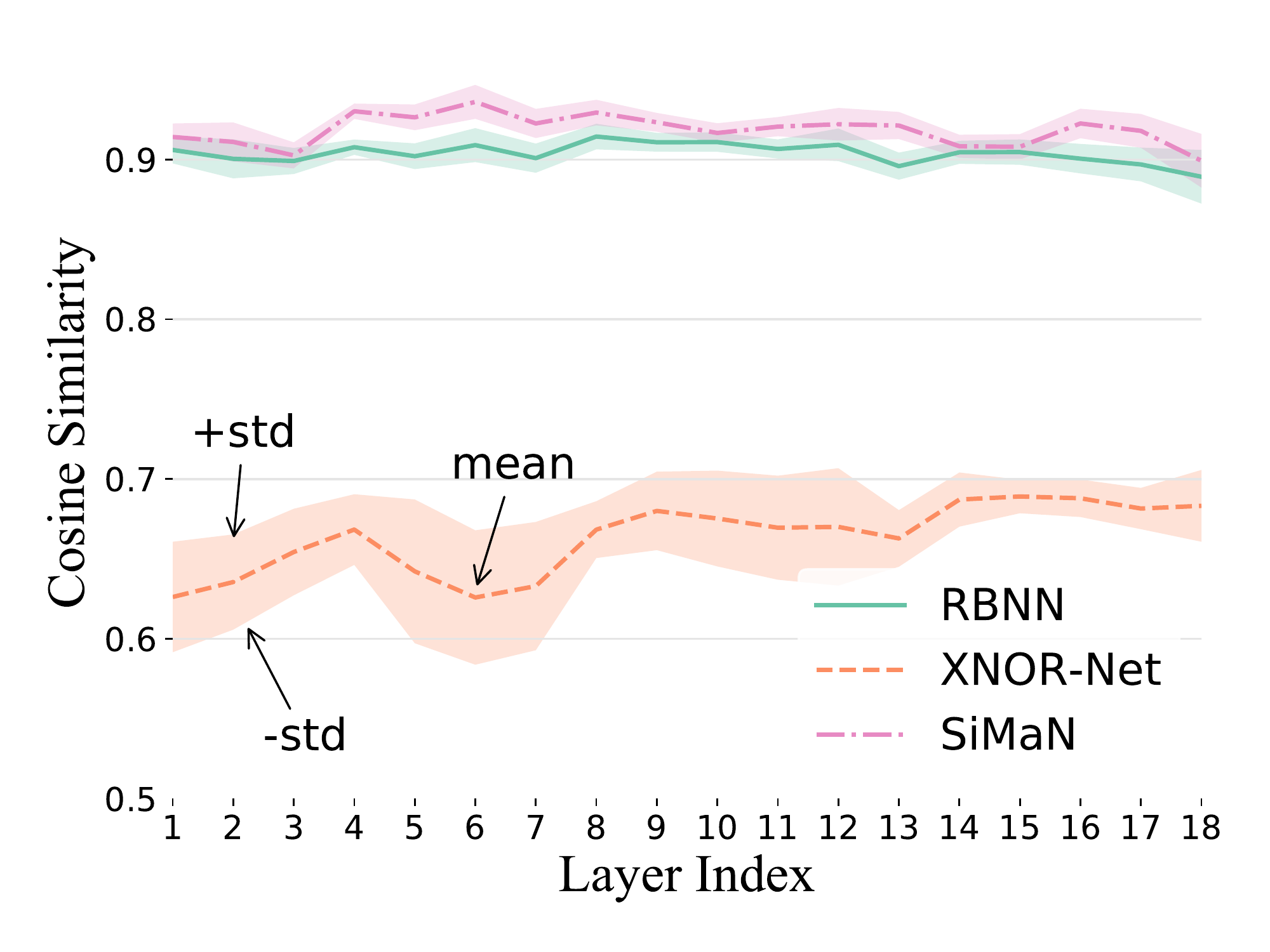}
    \caption{Cosine Similarity}
  \end{subfigure}
  \hspace{.3in}
  \vspace{.15in}
  \begin{subfigure}{0.43\linewidth}
    \includegraphics[width=\linewidth]{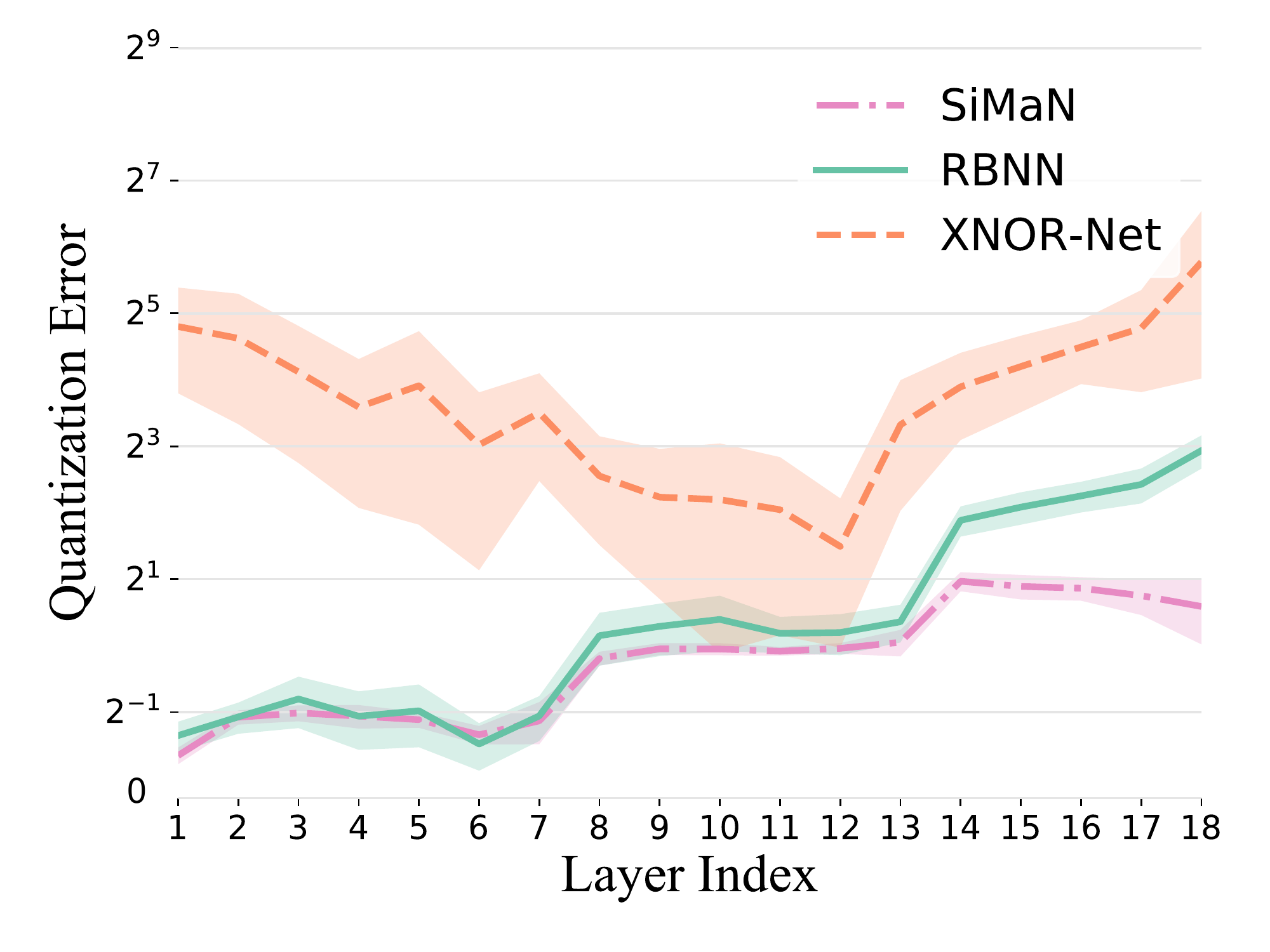}
    \caption{Quantization Error}
  \end{subfigure} 
  \vspace{-1.0em}
  \caption{Illustration of quantization error in various layers of ResNet-20.}
  \label{quantization_error}
  \vspace{-1.0em}
\end{figure*}

\subsection{Ablation Studies}\label{ablation_studies}

\begin{table}[!t]
\centering
\caption{Ablation studies with/without the $\ell_2$ regularization and half-half binarization (ResNet-18 on ImageNet).\label{ablation}} 
\vspace{-0.5em}
\begin{tabular}{ccccc}
\toprule
& $\ell_2$ regularization & half-half & Top-1(\%) & Top-5(\%)  \\\hline
SiMaN$_1$ &\Checkmark   &\XSolidBrush     &55.1   &75.5    \\\hline
SiMaN$_2$ &\XSolidBrush   &\XSolidBrush   &57.3   &77.4     \\\hline
SiMaN$_3$ &\Checkmark   &\Checkmark       &59.2  &81.5        \\\hline
SiMaN &\XSolidBrush   &\Checkmark      &60.1    &82.3        \\
\bottomrule
\end{tabular}
\end{table}

In this subsection, we conduct ablation studies of different variants to demonstrate the efficacy of our SiMaN, as well as quantization error to demonstrate the superiority of our analytical discrete optimization.

\textbf{SiMaN Variants}.
Our SiMaN is built by removing the $\ell_2$ regularization and enforcing the half-half strategy in Eq.\,(\ref{half-half}). To analyze their influence, in Table\,\ref{ablation}, we develop three variants, including 
(1) SiMaN$_1$: The $\ell_2$ regularization is added while removing the half-half binarization. This variant simply implements binarization based on the proof process of Corollary 1. It results in around $37\%$ of weights encoded into $+1$s, as analyzed in Corollary 2, which  fails to maximize the entropy information, and thus leads to poorer accuracies of 55.1\% for the top-1 and 75.5\% for the top-5.
(2) SiMaN$_2$: Both the $\ell_2$ regularization and the half-half binarization are removed. It shows better top-1 ($57.3\%$) and top-5 ($77.4\%$) accuracies  since the removal of the $\ell_2$ regularization breaks the Laplacian distribution and results in around $51\%$ of weights being encoded into $+1$s as experimentally verified in Fig.\,\ref{ratio}.
(3) SiMaN$_3$: Both the $\ell_2$ regularization and the half-half binarization are added. Though the performance increases, it is still limited. This is because, the half-half binarization with the $\ell_2$ regularization causes a large angle deviation of around $30.66^\circ - 36.35^\circ$ as analyzed in Sec.\,\ref{maximizing_bit_entropy}, from the optimal binarization of our learning objective in Eq.\,(\ref{bw}). 

Based on SiMaN$_3$, our SiMaN further removes the $\ell_2$ regularization. On one hand, this ensures the maximal bit entropy; on the other hand, it ensures that the half-half binarization closely matches the optimal binarization (only $7.97^\circ - 8.05^\circ$ angle deviation as analyzed in Sec.\,\ref{maximizing_bit_entropy}), thereby leading to the best performance in Table\,\ref{ablation}.


\textbf{Quantization Error}.
Recall in Sec.\,\ref{introduction}, to mitigate the quantization error, XNOR-Net~\cite{rastegari2016xnor} compensates for the norm gap between the full-precision weight and the corresponding binarization. Our conference implementation of RBNN~\cite{lin2020rotated} introduces a rotation matrix to reduce the angular bias. In this paper, we reformulate the angle alignment objective and derive an analytical discrete solution. 
To validate the theoretical claims on reducing quantization error, we measure the practical quantization errors across different layers of ResNet-20 as a toy example in Fig.\,\ref{quantization_error}. Similar observations can be found in other networks as well.

In Fig.\,\ref{quantization_error}(1), we first measure the cosine similarity between the full-precision weight vector and the corresponding binarization. We can see that our earlier implementation of RBNN achieves a significantly higher cosine similarity than XNOR-Net does, implying fewer angular bias. Upon RBNN, our SiMaN further increases the cosine similarity across different network layers.
Consequently, in Fig.\,\ref{quantization_error}(2), though XNOR-Net mitigates quantization error from the norm gap, quantization error still heavily accumulates in most layers since the angular bias remains unsolved. By weight rotation, our conference version of RBNN greatly closes the angular bias thus it can effectively decrease quantization error. Nevertheless, the alternating optimization in RBNN leads to sub-optimal binarization. Therefore, the quantization error in top layers still stays in a relatively large state. By contrast, the analytical optimization in SiMaN enables to further reduce the quantization error to a very small level, providing one new perspective of BNN optimization.

\subsection{Convergence}

\begin{figure}[!t]
\begin{center}
\includegraphics[width=1.0\linewidth]{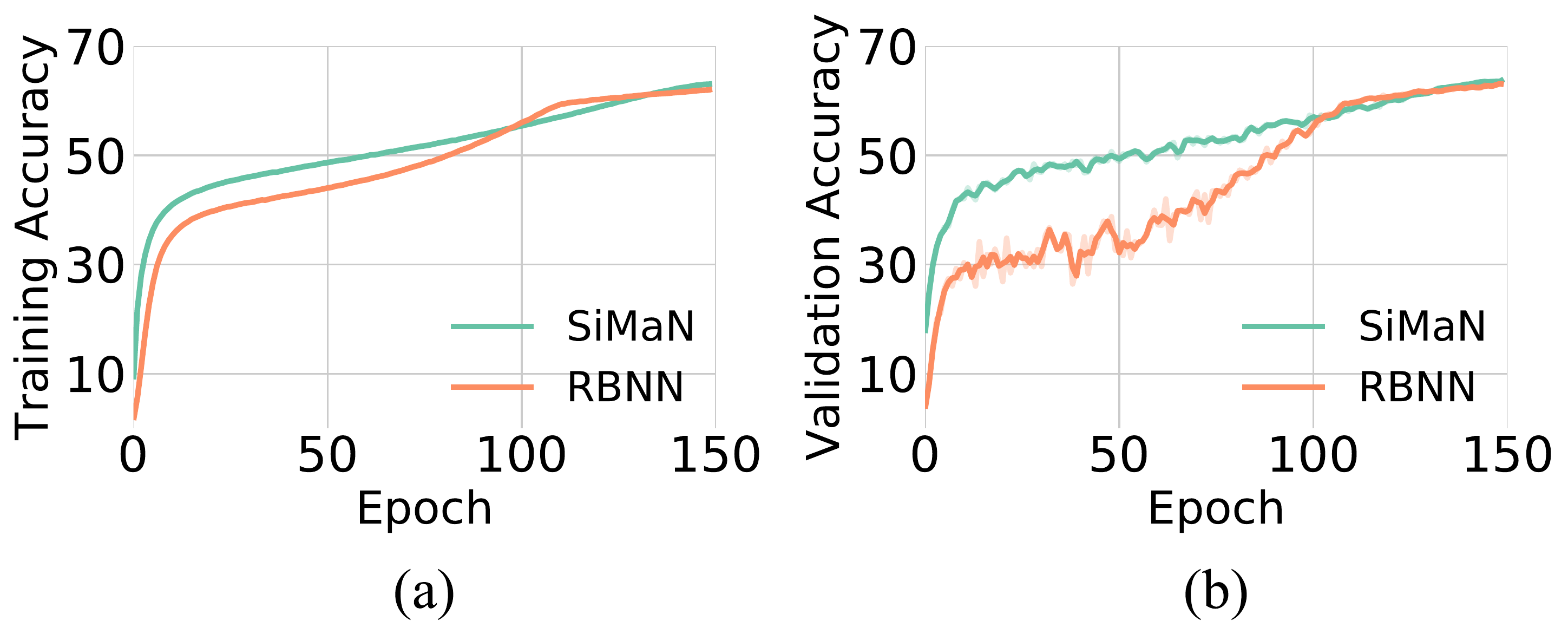}
\end{center}
\vspace{-0.8em}
   \caption{Comparison of training and validation accuracy curves between our SiMaN and RBNN (ResNet-34 on ImageNet).}
\label{convergence}
\vspace{-1.0em}
\end{figure}

We further show the convergence ability of our SiMaN, and compare with its conference version of RBNN which implements network binarization with the sign function~\cite{lin2020rotated}. The experiments in Fig.\,\ref{convergence} show that our sign-to-magnitude weight binarization has a significantly better ability to converge during BNN training than the traditional sign-based optimization on both training and validation sets, which demonstrates the feasibility of our discrete solution in learning BNNs.

\begin{table}[!t]
\centering
\caption{Comparison with the state-of-the-arts on CIFAR-10. W/A denotes the bit length of the weights and activations. Top-1 accuracy is reported.\label{cifar}} 
\vspace{-0.5em}
\begin{tabular}{cccc}
\toprule
Network                    & Method     & W/A   & \tabincell{c}{Top-1 (\%)} \\ \toprule
\multirow{5}{*}{\tabincell{c}{ResNet\\-18}} & Full-precision & 32/32 & 94.8\\\cline{2-4}
& RAD~\cite{ding2019regularizing}     & 1/1   & 90.5 \\\cline{2-4}
& IR-Net~\cite{qin2020forward}        & 1/1   & 91.5  \\\cline{2-4} 
& RBNN~\cite{lin2020rotated}          & 1/1   & 92.2  \\\cline{2-4}
& \textbf{SiMaN (Ours)}                  & 1/1   &\textbf{92.5} \\
\midrule
\multirow{6}{*}{\tabincell{c}{ResNet\\-20}} & Full-precision         & 32/32 & 92.1 \\\cline{2-4}
& DoReFa~\cite{zhou2016dorefa}     & 1/1   & 79.3 \\\cline{2-4}
& DSQ~\cite{gong2019differentiable}   & 1/1   & 84.1 \\\cline{2-4}
& SLB~\cite{yang2020searching}        & 1/1   & 85.5 \\\cline{2-4}
& LNS~\cite{han2020training}          &1/1    &85.8 \\\cline{2-4}
& IR-Net~\cite{qin2020forward}        & 1/1   & 86.5 \\\cline{2-4}
&RBNN~\cite{lin2020rotated}        & 1/1   & 87.8  \\\cline{2-4}
& \textbf{SiMaN (Ours)}                  & 1/1  &\textbf{87.4}  \\
\midrule
\multirow{10}{*}{\tabincell{c}{VGG\\-small}} & Full-precision        & 32/32 & 94.1 \\\cline{2-4}          
& XNOR-Net~\cite{rastegari2016xnor}& 1/1   & 89.8 \\\cline{2-4}
& BNN~\cite{courbariaux2016binarized}   & 1/1   & 89.9 \\ \cline{2-4}
& DoReFa~\cite{zhou2016dorefa}     & 1/1   & 90.2 \\\cline{2-4}
& RAD~\cite{ding2019regularizing}     & 1/1   & 90.0 \\\cline{2-4}
& DSQ~\cite{gong2019differentiable}   & 1/1   & 91.7 \\\cline{2-4}
& IR-Net~\cite{qin2020forward}        & 1/1   & 90.4 \\\cline{2-4}
& RBNN~\cite{lin2020rotated}          & 1/1   & 91.3 \\\cline{2-4}
& SLB~\cite{yang2020searching}        & 1/1   & 92.0 \\\cline{2-4}
& \textbf{SiMaN (Ours)}                & 1/1   &\textbf{92.5} \\
\bottomrule
\end{tabular}
\end{table}

\subsection{Results on CIFAR-10}

We first conduct detailed studies on CIFAR-10 for the proposed SiMaN as shown in Table\,\ref{cifar}. 
Despite that RBNN performs best in binarizing ResNet-20, our sign-to-magnitude binarization consistently outperforms the recent sign-based state-of-the-arts. Specifically, SiMaN outperforms RBNN~\cite{lin2020rotated} and SLB~\cite{yang2020searching} by $0.3\%$ and $0.5\%$ in binarizing ResNet-18 and VGG-small, respectively. 
The results emphasize the importance of building discrete optimization to pursue high-quality weight binarization. Importantly, compared with the conference version, \emph{i.e.}, RBNN, our SiMaN increases the performance to $92.5\%$ when binarizing VGG-small, leading to a performance gain of $1.2\%$. Besides, SiMaN also merits in its easy implementation, where the median of absolute weights acts as the boundary between $0$s and $+1$s to ensure the bit entropy maximization, while the angle alignment is also guaranteed, as analyzed in Sec.\,\ref{maximizing_bit_entropy}. In contrast, RBNN has to learn two complex rotation matrices and applies them in the beginning phase of each training epoch in order to reduce the angle bias.
Note that, for ResNet-20, we realize that smaller quantization error (see Fig.\,\ref{quantization_error}) does not necessarily lead to better performance. This indicates that there may exist an unexplored optimal solution that is not related to the quantization error. Nevertheless, it has been widely accepted in literature that quantization error can improve the BNN performance in most cases, which has been demonstrated by the experimental results (except ResNet-20) in this paper as well. This paper also addresses the quantization error problem.

\subsection{Results on ImageNet}

We also conduct similar  experiments on ImageNet to validate the performance of SiMaN on a large-scale dataset. Two common networks, ResNet-18 and ResNet-34, are adopted for binarization. Table\,\ref{imagenet} shows the results of SiMaN and several other binarization methods.
The performance of SiMaN on ImageNet also takes the leading place. Specifically, with ResNet-18, SiMaN achieves $60.1\%$ top-1 and $82.3\%$ top-5 accuracies, respectively, with $0.2\%$ and $0.4\%$ improvements over its conference version of RBNN. 
By doubling the number of residual blocks and shrinking per-block convolutions, ReActNet~\cite{liu2020reactnet} achieves 65.9\% top-1 accuracy. To manifest the advantage of our discrete optimization, we further perform SiMaN on the modified network structure and obtain a better performance of 66.1\%. 
With ResNet-34, it achieves a top-1 accuracy of $63.9\%$ and a top-5 accuracy of $84.8\%$, outperforming  RBNN by $0.8\%$ and $0.4\%$, respectively.

The performance improvements in Table\,\ref{cifar} and Table\,\ref{imagenet} strongly demonstrate the impact of exploring discrete optimization and the effectiveness of our magnitude-based discrete solution in constructing a high-performing BNN.

\begin{table}[!t]
\centering
\caption{Comparison with the state-of-the-arts on ImageNet. W/A denotes the bit length of the weights and activations. Both top-1 and top-5 accuracies are reported. SiMaN$^{\dag}$ means using the same network and training setting as ReActNet~\cite{liu2020reactnet}.\label{imagenet}} 
\vspace{-0.5em}
\begin{tabular}{ccccc}
\toprule
Network                   & Method & W/A & \tabincell{c}{Top-1 (\%)} & \tabincell{c}{Top-5 (\%)} \\ 
\toprule
\multirow{9}{*}{\tabincell{c}{ResNet\\-18}} & Full-precision        &32/32  & 69.6    & 89.2 \\\cline{2-5}
& BNN~\cite{courbariaux2016binarized}   & 1/1   & 42.2    & 67.1 \\\cline{2-5}
& XNOR-Net~\cite{rastegari2016xnor}& 1/1   & 51.2   & 73.2 \\\cline{2-5}
& DoReFa~\cite{zhou2016dorefa}     & 1/2   & 53.4    & -      \\\cline{2-5}
& HWGQ~\cite{cai2017deep}             & 1/2   & 59.6    & 82.2 \\\cline{2-5}
& TBN~\cite{wan2018tbn}               & 1/2   & 55.6    & 79.0 \\\cline{2-5}
& Bi-Real~\cite{liu2018bi}          & 1/1   & 56.4    & 79.5 \\\cline{2-5}
& PDNN~\cite{gu2019projection}        & 1/1   & 57.3    & 80.0 \\\cline{2-5}
& BONN~\cite{gu2019bayesian}          & 1/1   & 59.3    & 81.6 \\\cline{2-5}
& Si-BNN~\cite{wang2020sparsity}   &1/1    &59.7 &81.8 \\\cline{2-5}
& IR-Net~\cite{qin2020forward}        & 1/1   & 58.1    & 80.0 \\\cline{2-5}
& LNS~\cite{han2020training}           &1/1 &59.4 &81.7  \\\cline{2-5}
& RBNN~\cite{lin2020rotated}          & 1/1   & 59.9    & 81.9 \\\cline{2-5}
& \textbf{SiMaN (Ours)}                    & 1/1   &\textbf{60.1}   &\textbf{82.3} \\\cline{2-5}
&ReActNet~\cite{liu2020reactnet} &1/1 &65.9 &- \\\cline{2-5}
&\textbf{SiMaN}$^{\dag}$ &1/1 &\textbf{66.1} &\textbf{85.9} \\
\midrule
\multirow{6}{*}{\tabincell{c}{ResNet\\-34}} & Full-precision         &32/32 & 73.3     & 91.3 \\\cline{2-5}
& ABC-Net~\cite{lin2017towards}       & 1/1  & 52.4     & 76.5\\\cline{2-5}
& Bi-Real~\cite{liu2018bi}          & 1/1  & 62.2    & 83.9\\ \cline{2-5}
& IR-Net \cite{qin2020forward}        & 1/1  & 62.9     & 84.1\\\cline{2-5}
& RBNN~\cite{lin2020rotated}          & 1/1  & 63.1     & 84.4\\ \cline{2-5}
&\textbf{SiMaN (Ours)}              & 1/1  &\textbf{63.9}  &\textbf{84.8}\\
\bottomrule
\end{tabular}
\end{table}

\section{Conclusion}\label{conclusion}

In this paper, we proposed a novel sign-to-magnitude network binarization (SiMaN) scheme that avoids the dependency on the sign function, to optimize a binary neural network for higher accuracy. Our SiMaN reformulates the angle alignment between the weight vector and its binarization as being constrained to $\{0, +1\}$. We proved that an analytical discrete solution can be attained in a computationally efficient manner by encoding into $+1$s the high-magnitude weights, and $0$s otherwise. We also mathematically proved that the learned weights roughly follow a Laplacian distribution, which is harmful to bit entropy maximization. To address the problem, we have shown that simply removing the $\ell_2$ regularization during network training can break  the Laplacian distribution and lead to a half-half distribution of binarized weights. As a result, the complexity of our binarization could be further simplified by encoding into $+1$ weights within the largest top-half magnitude, and $0$ otherwise. Our experimental results demonstrate the significant performance improvement of SiMaN.

\section*{Acknowledgement}
This work was supported by the National Science Fund for Distinguished Young Scholars (No.62025603), the National Natural Science Foundation of China (No. U21B2037, No. 62176222, No. 62176223, No. 62176226, No. 62072386, No. 62072387, No. 62072389, and No. 62002305), Guangdong Basic and Applied Basic Research Foundation (No.2019B1515120049), and the Natural Science Foundation of Fujian Province of China (No.2021J01002).

\bibliographystyle{IEEEtran}
\bibliography{mybib.bib}

\begin{thebibliography}{10}
\providecommand{\url}[1]{#1}
\csname url@samestyle\endcsname
\providecommand{\newblock}{\relax}
\providecommand{\bibinfo}[2]{#2}
\providecommand{\BIBentrySTDinterwordspacing}{\spaceskip=0pt\relax}
\providecommand{\BIBentryALTinterwordstretchfactor}{4}
\providecommand{\BIBentryALTinterwordspacing}{\spaceskip=\fontdimen2\font plus
\BIBentryALTinterwordstretchfactor\fontdimen3\font minus
  \fontdimen4\font\relax}
\providecommand{\BIBforeignlanguage}[2]{{%
\expandafter\ifx\csname l@#1\endcsname\relax
\typeout{** WARNING: IEEEtran.bst: No hyphenation pattern has been}%
\typeout{** loaded for the language `#1'. Using the pattern for}%
\typeout{** the default language instead.}%
\else
\language=\csname l@#1\endcsname
\fi
#2}}
\providecommand{\BIBdecl}{\relax}
\BIBdecl

\bibitem{he2016deep}
K.~He, X.~Zhang, S.~Ren, and J.~Sun, ``Deep residual learning for image
  recognition,'' in \emph{Proceedings of the IEEE Conference on Computer Vision
  and Pattern Recognition (CVPR)}, 2016, pp. 770--778.

\bibitem{wei2015hcp}
Y.~Wei, W.~Xia, M.~Lin, J.~Huang, B.~Ni, J.~Dong, Y.~Zhao, and S.~Yan, ``Hcp: A
  flexible cnn framework for multi-label image classification,'' \emph{IEEE
  Transactions on Pattern Analysis and Machine Intelligence (T-PAMI)}, vol.~38,
  no.~9, pp. 1901--1907, 2015.

\bibitem{he2015spatial}
K.~He, X.~Zhang, S.~Ren, and J.~Sun, ``Spatial pyramid pooling in deep
  convolutional networks for visual recognition,'' \emph{IEEE Transactions on
  Pattern Analysis and Machine Intelligence (T-PAMI)}, vol.~37, no.~9, pp.
  1904--1916, 2015.

\bibitem{redmon2016you}
J.~Redmon, S.~Divvala, R.~Girshick, and A.~Farhadi, ``You only look once:
  Unified, real-time object detection,'' in \emph{Proceedings of the IEEE
  Conference on Computer Vision and Pattern Recognition (CVPR)}, 2016, pp.
  779--788.

\bibitem{ren2016object}
S.~Ren, K.~He, R.~Girshick, X.~Zhang, and J.~Sun, ``Object detection networks
  on convolutional feature maps,'' \emph{IEEE Transactions on Pattern Analysis
  and Machine Intelligence (T-PAMI)}, vol.~39, no.~7, pp. 1476--1481, 2016.

\bibitem{zhang2021learning}
X.~Zhang, F.~Wan, C.~Liu, X.~Ji, and Q.~Ye, ``Learning to match anchors for
  visual object detection,'' \emph{IEEE Transactions on Pattern Analysis and
  Machine Intelligence (T-PAMI)}, vol.~44, no.~6, pp. 3096--3109, 2021.

\bibitem{long2015fully}
J.~Long, E.~Shelhamer, and T.~Darrell, ``Fully convolutional networks for
  semantic segmentation,'' in \emph{Proceedings of the IEEE Conference on
  Computer Vision and Pattern Recognition (CVPR)}, 2015, pp. 3431--3440.

\bibitem{badrinarayanan2017segnet}
V.~Badrinarayanan, A.~Kendall, and R.~Cipolla, ``Segnet: A deep convolutional
  encoder-decoder architecture for image segmentation,'' \emph{IEEE
  Transactions on Pattern Analysis and Machine Intelligence (T-PAMI)}, vol.~39,
  no.~12, pp. 2481--2495, 2017.

\bibitem{shelhamer2017fully}
E.~Shelhamer, J.~Long, and T.~Darrell, ``Fully convolutional networks for
  semantic segmentation,'' \emph{IEEE Transactions on Pattern Analysis and
  Machine Intelligence (T-PAMI)}, vol.~39, no.~4, pp. 640--651, 2017.

\bibitem{han2015learning}
S.~Han, J.~Pool, J.~Tran, and W.~Dally, ``Learning both weights and connections
  for efficient neural network,'' in \emph{Proceedings of the Advances in
  Neural Information Processing Systems (NeurIPS)}, 2015, pp. 1135--1143.

\bibitem{frankle2019lottery}
J.~Frankle and M.~Carbin, ``The lottery ticket hypothesis: Finding sparse,
  trainable neural networks,'' in \emph{Proceedings of the International
  Conference on Learning Representations (ICLR)}, 2019.

\bibitem{luo2018thinet}
J.-H. Luo, H.~Zhang, H.-Y. Zhou, C.-W. Xie, J.~Wu, and W.~Lin, ``Thinet:
  pruning cnn filters for a thinner net,'' \emph{IEEE Transactions on Pattern
  Analysis and Machine Intelligence (T-PAMI)}, vol.~41, no.~10, pp. 2525--2538,
  2018.

\bibitem{lin2020hrank}
M.~Lin, R.~Ji, Y.~Wang, Y.~Zhang, B.~Zhang, Y.~Tian, and L.~Shao, ``Hrank:
  Filter pruning using high-rank feature map,'' in \emph{Proceedings of the
  IEEE Conference on Computer Vision and Pattern Recognition (CVPR)}, 2020, pp.
  1529--1538.

\bibitem{zhang2018shufflenet}
X.~Zhang, X.~Zhou, M.~Lin, and J.~Sun, ``Shufflenet: An extremely efficient
  convolutional neural network for mobile devices,'' in \emph{Proceedings of
  the IEEE Conference on Computer Vision and Pattern Recognition (CVPR)}, 2018,
  pp. 6848--6856.

\bibitem{ma2018shufflenet}
N.~Ma, X.~Zhang, H.-T. Zheng, and J.~Sun, ``Shufflenet v2: Practical guidelines
  for efficient cnn architecture design,'' in \emph{Proceedings of the European
  Conference on Computer Vision (ECCV)}, 2018, pp. 116--131.

\bibitem{howard2017mobilenets}
A.~G. Howard, M.~Zhu, B.~Chen, D.~Kalenichenko, W.~Wang, T.~Weyand,
  M.~Andreetto, and H.~Adam, ``Mobilenets: Efficient convolutional neural
  networks for mobile vision applications,'' \emph{arXiv preprint
  arXiv:1704.04861}, 2017.

\bibitem{sandler2018mobilenetv2}
M.~Sandler, A.~Howard, M.~Zhu, A.~Zhmoginov, and L.-C. Chen, ``Mobilenetv2:
  Inverted residuals and linear bottlenecks,'' in \emph{Proceedings of the IEEE
  Conference on Computer Vision and Pattern Recognition (CVPR)}, 2018, pp.
  4510--4520.

\bibitem{howard2019searching}
A.~Howard, M.~Sandler, G.~Chu, L.-C. Chen, B.~Chen, M.~Tan, W.~Wang, Y.~Zhu,
  R.~Pang, V.~Vasudevan \emph{et~al.}, ``Searching for mobilenetv3,'' in
  \emph{Proceedings of the IEEE Conference on Computer Vision and Pattern
  Recognition (CVPR)}, 2019, pp. 1314--1324.

\bibitem{han2020ghostnet}
K.~Han, Y.~Wang, Q.~Tian, J.~Guo, C.~Xu, and C.~Xu, ``Ghostnet: More features
  from cheap operations,'' in \emph{Proceedings of the IEEE Conference on
  Computer Vision and Pattern Recognition (CVPR)}, 2020, pp. 1580--1589.

\bibitem{lin2018holistic}
S.~Lin, R.~Ji, C.~Chen, D.~Tao, and J.~Luo, ``Holistic cnn compression via
  low-rank decomposition with knowledge transfer,'' \emph{IEEE Transactions on
  Pattern Analysis and Machine Intelligence (T-PAMI)}, vol.~41, no.~12, pp.
  2889--2905, 2018.

\bibitem{hayashi2019exploring}
K.~Hayashi, T.~Yamaguchi, Y.~Sugawara, and S.-i. Maeda, ``Exploring unexplored
  tensor network decompositions for convolutional neural networks,'' in
  \emph{Proceedings of the Advances in Neural Information Processing Systems
  (NeurIPS)}, 2019, pp. 5552--5562.

\bibitem{cai2017deep}
Z.~Cai, X.~He, J.~Sun, and N.~Vasconcelos, ``Deep learning with low precision
  by half-wave gaussian quantization,'' in \emph{Proceedings of the IEEE
  Conference on Computer Vision and Pattern Recognition (CVPR)}, 2017, pp.
  5918--5926.

\bibitem{han2020training}
K.~Han, Y.~Wang, Y.~Xu, C.~Xu, E.~Wu, and C.~Xu, ``Training binary neural
  networks through learning with noisy supervision,'' in \emph{Proceedings of
  the International Conference on Machine Learning (ICML)}, 2020, pp.
  4017--4026.

\bibitem{lin2020rotated}
M.~Lin, R.~Ji, Z.~Xu, B.~Zhang, Y.~Wang, Y.~Wu, F.~Huang, and C.-W. Lin,
  ``Rotated binary neural network,'' in \emph{Proceedings of the Advances in
  Neural Information Processing Systems (NeurIPS)}, 2020, pp. 7474--7485.

\bibitem{rastegari2016xnor}
M.~Rastegari, V.~Ordonez, J.~Redmon, and A.~Farhadi, ``Xnor-net: Imagenet
  classification using binary convolutional neural networks,'' in
  \emph{Proceedings of the European Conference on Computer Vision (ECCV)},
  2016, pp. 525--542.

\bibitem{deng2009imagenet}
J.~Deng, W.~Dong, R.~Socher, L.-J. Li, K.~Li, and L.~Fei-Fei, ``Imagenet: A
  large-scale hierarchical image database,'' in \emph{Proceedings of the IEEE
  Conference on Computer Vision and Pattern Recognition (CVPR)}, 2009, pp.
  248--255.

\bibitem{lin2017towards}
X.~Lin, C.~Zhao, and W.~Pan, ``Towards accurate binary convolutional neural
  network,'' in \emph{Proceedings of the Advances in Neural Information
  Processing Systems (NeurIPS)}, 2017, pp. 345--353.

\bibitem{liu2018bi}
Z.~Liu, B.~Wu, W.~Luo, X.~Yang, W.~Liu, and K.-T. Cheng, ``Bi-real net:
  Enhancing the performance of 1-bit cnns with improved representational
  capability and advanced training algorithm,'' in \emph{Proceedings of the
  European Conference on Computer Vision (ECCV)}, 2018, pp. 722--737.

\bibitem{qin2020forward}
H.~Qin, R.~Gong, X.~Liu, M.~Shen, Z.~Wei, F.~Yu, and J.~Song, ``Forward and
  backward information retention for accurate binary neural networks,'' in
  \emph{Proceedings of the IEEE Conference on Computer Vision and Pattern
  Recognition (CVPR)}, 2020, pp. 2250--2259.

\bibitem{gu2019bayesian}
J.~Gu, J.~Zhao, X.~Jiang, B.~Zhang, J.~Liu, G.~Guo, and R.~Ji, ``Bayesian
  optimized 1-bit cnns,'' in \emph{Proceedings of the IEEE International
  Conference on Computer Vision (ICCV)}, 2019, pp. 4909--4917.

\bibitem{courbariaux2015binaryconnect}
M.~Courbariaux, Y.~Bengio, and J.-P. David, ``Binaryconnect: Training deep
  neural networks with binary weights during propagations,'' in
  \emph{Proceedings of the Advances in Neural Information Processing Systems
  (NeurIPS)}, 2015, pp. 3123--3131.

\bibitem{courbariaux2016binarized}
M.~Courbariaux, I.~Hubara, D.~Soudry, R.~El-Yaniv, and Y.~Bengio, ``Binarized
  neural networks: Training deep neural networks with weights and activations
  constrained to+ 1 or-1,'' \emph{arXiv preprint arXiv:1602.02830}, 2016.

\bibitem{martinez2019training}
B.~Martinez, J.~Yang, A.~Bulat, and G.~Tzimiropoulos, ``Training binary neural
  networks with real-to-binary convolutions,'' in \emph{International
  Conference on Learning Representations (ICLR)}, 2020.

\bibitem{krizhevsky2009learning}
A.~Krizhevsky, ``Learning multiple layers of features from tiny images,''
  \emph{Master's thesis, University of Tront}, 2009.

\bibitem{bengio2013estimating}
Y.~Bengio, N.~L{\'e}onard, and A.~Courville, ``Estimating or propagating
  gradients through stochastic neurons for conditional computation,''
  \emph{arXiv preprint arXiv:1308.3432}, 2013.

\bibitem{simons2019review}
T.~Simons and D.-J. Lee, ``A review of binarized neural networks,''
  \emph{Electronics}, vol.~8, no.~6, p. 661, 2019.

\bibitem{qin2020binary}
H.~Qin, R.~Gong, X.~Liu, X.~Bai, J.~Song, and N.~Sebe, ``Binary neural
  networks: A survey,'' \emph{Pattern Recognition (PR)}, vol. 105, p. 107281,
  2020.

\bibitem{bulat2019xnor}
A.~Bulat and G.~Tzimiropoulos, ``Xnor-net++: Improved binary neural networks,''
  \emph{arXiv preprint arXiv:1909.13863}, 2019.

\bibitem{xu2021recu}
Z.~Xu, M.~Lin, J.~Liu, J.~Chen, L.~Shao, Y.~Gao, Y.~Tian, and R.~Ji, ``Recu:
  Reviving the dead weights in binary neural networks,'' in \emph{Proceedings
  of the IEEE International Conference on Computer Vision (ICCV)}, 2021, pp.
  5198--5208.

\bibitem{darabi2018bnn+}
S.~Darabi, M.~Belbahri, M.~Courbariaux, and V.~P. Nia, ``Bnn+: Improved binary
  network training,'' \emph{arXiv preprint arXiv:1812.11800}, 2018.

\bibitem{xu2021learning}
Y.~Xu, K.~Han, C.~Xu, Y.~Tang, C.~Xu, and Y.~Wang, ``Learning frequency domain
  approximation for binary neural networks,'' in \emph{Proceedings of the
  Advances in Neural Information Processing Systems (NeurIPS)}, 2021, pp.
  25\,553--25\,565.

\bibitem{peters2018probabilistic}
J.~W. Peters and M.~Welling, ``Probabilistic binary neural networks,''
  \emph{arXiv preprint arXiv:1809.03368}, 2018.

\bibitem{shayer2018learning}
O.~Shayer, D.~Levi, and E.~Fetaya, ``Learning discrete weights using the local
  reparameterization trick,'' in \emph{Proceedings of the International
  Conference on Learning Representations (ICLR)}, 2018.

\bibitem{qin2020bipointnet}
H.~Qin, Z.~Cai, M.~Zhang, Y.~Ding, H.~Zhao, S.~Yi, X.~Liu, and H.~Su,
  ``Bipointnet: Binary neural network for point clouds,'' in \emph{Proceedings
  of the International Conference on Learning Representations (ICLR)}, 2022.

\bibitem{leng2018extremely}
C.~Leng, Z.~Dou, H.~Li, S.~Zhu, and R.~Jin, ``Extremely low bit neural network:
  Squeeze the last bit out with admm,'' in \emph{Proceedings of the AAAI
  Conference on Artificial Intelligence (AAAI)}, 2018, pp. 3466--3473.

\bibitem{alizadeh2018empirical}
M.~Alizadeh, J.~Fern{\'a}ndez-Marqu{\'e}s, N.~D. Lane, and Y.~Gal, ``An
  empirical study of binary neural networks' optimisation,'' in
  \emph{Proceedings of the International Conference on Learning Representations
  (ICLR)}, 2018.

\bibitem{bethge2019back}
J.~Bethge, H.~Yang, M.~Bornstein, and C.~Meinel, ``Back to simplicity: How to
  train accurate bnns from scratch?'' \emph{arXiv preprint arXiv:1906.08637},
  2019.

\bibitem{helwegen2019latent}
K.~Helwegen, J.~Widdicombe, L.~Geiger, Z.~Liu, K.-T. Cheng, and R.~Nusselder,
  ``Latent weights do not exist: Rethinking binarized neural network
  optimization,'' in \emph{Proceedings of the Advances in Neural Information
  Processing Systems (NeurIPS)}, 2019, pp. 7533--7544.

\bibitem{anderson2018high}
A.~G. Anderson and C.~P. Berg, ``The high-dimensional geometry of binary neural
  networks,'' in \emph{International Conference on Learning Representations
  (ICLR)}, 2018.

\bibitem{wang2021sub}
Y.~Wang, Y.~Yang, F.~Sun, and A.~Yao, ``Sub-bit neural networks: Learning to
  compress and accelerate binary neural networks,'' in \emph{Proceedings of the
  IEEE International Conference on Computer Vision (ICCV)}, 2021, pp.
  5360--5369.

\bibitem{dockhorn2021demystifying}
T.~Dockhorn, Y.~Yu, E.~Sari, M.~Zolnouri, and V.~Partovi~Nia, ``Demystifying
  and generalizing binaryconnect,'' in \emph{Proceedings of the Advances in
  Neural Information Processing Systems (NeurIPS)}, 2021, pp. 13\,202--13\,216.

\bibitem{gu2019projection}
J.~Gu, C.~Li, B.~Zhang, J.~Han, X.~Cao, J.~Liu, and D.~Doermann, ``Projection
  convolutional neural networks for 1-bit cnns via discrete back propagation,''
  in \emph{Proceedings of the AAAI Conference on Artificial Intelligence
  (AAAI)}, 2019, pp. 8344--8351.

\bibitem{ding2019regularizing}
R.~Ding, T.-W. Chin, Z.~Liu, and D.~Marculescu, ``Regularizing activation
  distribution for training binarized deep networks,'' in \emph{Proceedings of
  the IEEE Conference on Computer Vision and Pattern Recognition (CVPR)}, 2019,
  pp. 11\,408--11\,417.

\bibitem{hu2022elastic}
J.~Hu, W.~Ziheng, V.~Tan, Z.~Lu, M.~Zeng, and E.~Wu, ``Elastic-link for
  binarized neural network,'' \emph{arXiv preprint arXiv:2112.10149}, 2021.

\bibitem{liu2020reactnet}
Z.~Liu, Z.~Shen, M.~Savvides, and K.-T. Cheng, ``Reactnet: Towards precise
  binary neural network with generalized activation functions,'' in
  \emph{Proceedings of the European Conference on Computer Vision (ECCV)},
  2020, pp. 143--159.

\bibitem{hubara2016binarized}
I.~Hubara, M.~Courbariaux, D.~Soudry, R.~El-Yaniv, and Y.~Bengio, ``Binarized
  neural networks,'' in \emph{Proceedings of the Advances in Neural Information
  Processing Systems (NeurIPS)}, 2016, pp. 4107--4115.

\bibitem{zhou2016dorefa}
S.~Zhou, Y.~Wu, Z.~Ni, X.~Zhou, H.~Wen, and Y.~Zou, ``Dorefa-net: Training low
  bitwidth convolutional neural networks with low bitwidth gradients,''
  \emph{arXiv preprint arXiv:1606.06160}, 2016.

\bibitem{conforti2014integer}
M.~Conforti, G.~Cornu{\'e}jols, G.~Zambelli \emph{et~al.}, \emph{Integer
  programming}.\hskip 1em plus 0.5em minus 0.4em\relax Springer, 2014, vol.
  271.

\bibitem{banner2018post}
R.~Banner, Y.~Nahshan, E.~Hoffer, and D.~Soudry, ``Post-training 4-bit
  quantization of convolution networks for rapid-deployment,'' \emph{arXiv
  preprint arXiv:1810.05723}, 2018.

\bibitem{zhong2020towards}
K.~Zhong, T.~Zhao, X.~Ning, S.~Zeng, K.~Guo, Y.~Wang, and H.~Yang, ``Towards
  lower bit multiplication for convolutional neural network training,''
  \emph{arXiv preprint arXiv:2006.02804}, 2020.

\bibitem{shakarji2013theory}
C.~M. Shakarji and V.~Srinivasan, ``Theory and algorithms for weighted total
  least-squares fitting of lines, planes, and parallel planes to support
  tolerancing standards,'' \emph{Journal of Computing and Information Science
  in Engineering}, vol.~13, no.~3, 2013.

\bibitem{andrews1998special}
L.~C. Andrews, \emph{Special functions of mathematics for engineers}.\hskip 1em
  plus 0.5em minus 0.4em\relax Spie Press, 1998, vol.~49.

\bibitem{wan2018tbn}
D.~Wan, F.~Shen, L.~Liu, F.~Zhu, J.~Qin, L.~Shao, and H.~Tao~Shen, ``Tbn:
  Convolutional neural network with ternary inputs and binary weights,'' in
  \emph{Proceedings of the European Conference on Computer Vision (ECCV)},
  2018, pp. 315--332.

\bibitem{gong2019differentiable}
R.~Gong, X.~Liu, S.~Jiang, T.~Li, P.~Hu, J.~Lin, F.~Yu, and J.~Yan,
  ``Differentiable soft quantization: Bridging full-precision and low-bit
  neural networks,'' in \emph{Proceedings of the IEEE International Conference
  on Computer Vision (ICCV)}, 2019, pp. 4852--4861.

\bibitem{yang2020searching}
Z.~Yang, Y.~Wang, K.~Han, C.~Xu, C.~Xu, D.~Tao, and C.~Xu, ``Searching for
  low-bit weights in quantized neural networks,'' in \emph{Proceedings of the
  Advances in Neural Information Processing Systems (NeurIPS)}, 2020, pp.
  4091--4102.

\bibitem{wang2020sparsity}
P.~Wang, X.~He, G.~Li, T.~Zhao, and J.~Cheng, ``Sparsity-inducing binarized
  neural networks.'' in \emph{Proceedings of the AAAI Conference on Artificial
  Intelligence (AAAI)}, 2020, pp. 12\,192--12\,199.

\bibitem{zhang2018lq}
D.~Zhang, J.~Yang, D.~Ye, and G.~Hua, ``Lq-nets: Learned quantization for
  highly accurate and compact deep neural networks,'' in \emph{Proceedings of
  the European Conference on Computer Vision (ECCV)}, 2018, pp. 365--382.

\bibitem{paszke2019pytorch}
A.~Paszke, S.~Gross, F.~Massa, A.~Lerer, J.~Bradbury, G.~Chanan, T.~Killeen,
  Z.~Lin, N.~Gimelshein, L.~Antiga \emph{et~al.}, ``Pytorch: An imperative
  style, high-performance deep learning library,'' in \emph{Proceedings of the
  Advances in Neural Information Processing Systems (NeurIPS)}, 2019, pp.
  8026--8037.

\bibitem{hou2016loss}
L.~Hou, Q.~Yao, and J.~T. Kwok, ``Loss-aware binarization of deep networks,''
  in \emph{Proceedings of the International Conference on Learning
  Representations (ICLR)}, 2016.

\bibitem{wang2019learning}
Z.~Wang, J.~Lu, C.~Tao, J.~Zhou, and Q.~Tian, ``Learning channel-wise
  interactions for binary convolutional neural networks,'' in \emph{Proceedings
  of the IEEE Conference on Computer Vision and Pattern Recognition (CVPR)},
  2019, pp. 568--577.

\bibitem{bethge2020meliusnet}
J.~Bethge, C.~Bartz, H.~Yang, Y.~Chen, and C.~Meinel, ``Meliusnet: Can binary
  neural networks achieve mobilenet-level accuracy?'' \emph{arXiv preprint
  arXiv:2001.05936}, 2020.

\bibitem{zhu2019binary}
S.~Zhu, X.~Dong, and H.~Su, ``Binary ensemble neural network: More bits per
  network or more networks per bit?'' in \emph{Proceedings of the IEEE
  Conference on Computer Vision and Pattern Recognition (CVPR)}, 2019, pp.
  4923--4932.

\end{thebibliography}

\ifCLASSOPTIONcaptionsoff
  \newpage
\fi

\begin{IEEEbiography}[{\includegraphics[width=1in,height=1.25in,clip,keepaspectratio]{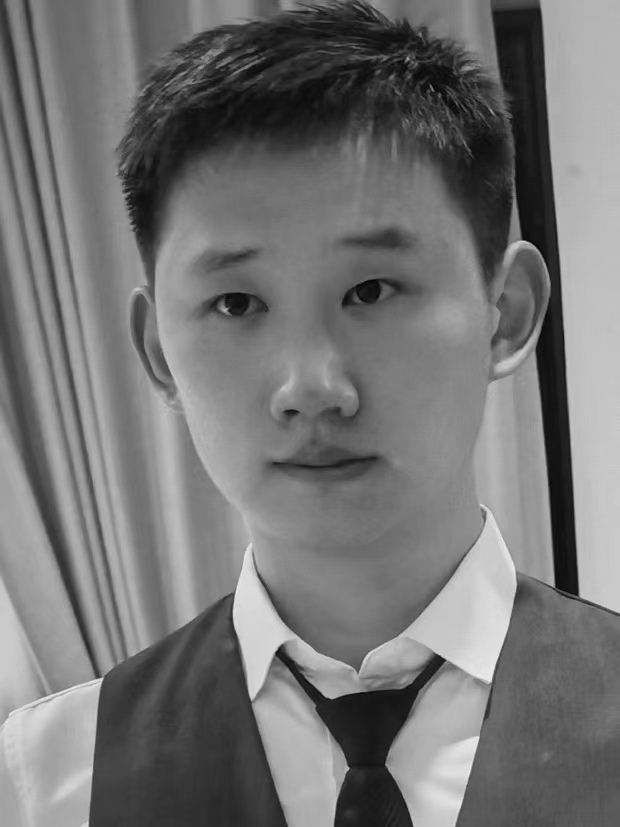}}]{Mingbao Lin} finished his M.S.-Ph.D. study and obtained the Ph.D. degree in intelligence science and technology from Xiamen University, Xiamen, China, in 2022. Earlier, he received the B.S. degree from Fuzhou University, Fuzhou, China, in 2016.

He is currently a senior researcher with the Tencent Youtu Lab, Shanghai, China. He has published over ten papers as the first author in top-tier journals and conferences, including IEEE TPAMI, IJCV, IEEE TIP, IEEE TNNLS, CVPR, NeurIPS, AAAI, IJCAI, ACM MM and so on. His current research interest includes network compression \& acceleration, and information retrieval.
\end{IEEEbiography}

\begin{IEEEbiography}[{\includegraphics[width=1in,height=1.25in,clip,keepaspectratio]{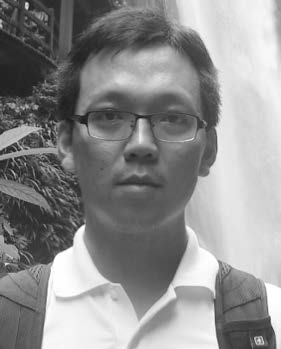}}]{Rongrong Ji}
(Senior Member, IEEE) is currently a Professor and the Director of the Intelligent Multimedia Technology Laboratory, and the Dean Assistant with the School of Information Science and Engineering, Xiamen University, Xiamen, China. His work mainly focuses on innovative technologies for multimedia signal processing, computer vision, and pattern recognition, with over 100 papers published in international journals and conferences. He is a member of the ACM. He was a recipient of the ACM Multimedia Best Paper Award and the Best Thesis Award of Harbin Institute of Technology. He serves as an Associate/Guest Editor for international journals and magazines such as \emph{Neurocomputing}, \emph{Signal Processing}, \emph{Multimedia Tools and Applications}, the \emph{IEEE Multimedia Magazine}, and the \emph{Multimedia Systems}. He also serves as program committee member for several Tier-$1$ international conferences.
\end{IEEEbiography}

\begin{IEEEbiography}[{\includegraphics[width=1in,height=1.25in,clip,keepaspectratio]{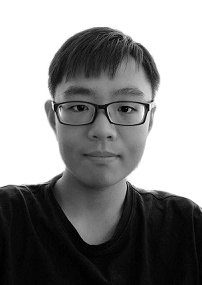}}]{Zihan Xu} received the B.S. degree in applied mathematics from Zhengzhou University, China, in 2019. He is currently pursuing the M.S. degree with Xiamen University, China. His research interests include computer vision and machine learning.
\end{IEEEbiography}

\begin{IEEEbiography}[{\includegraphics[width=1in,height=1.25in,clip,keepaspectratio]{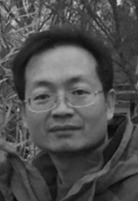}}]{Baochang Zhang}
(Senior Member, IEEE) received the B.S., M.S., and Ph.D. degrees in computer science from the Harbin Institute of the Technology, Harbin, China, in 1999, 2001, and 2006, respectively. From 2006 to 2008, he was a Research Fellow with The Chinese University of Hong Kong, Hong Kong, and also with Griffith University, Brisban, Australia. He is a researcher with the Zhongguancun Lab, Beijing, China. His current research interests include pattern recognition, machine learning, face recognition, and wavelets.
\end{IEEEbiography}

\begin{IEEEbiography}[{\includegraphics[width=1in,height=1.25in,clip,keepaspectratio]{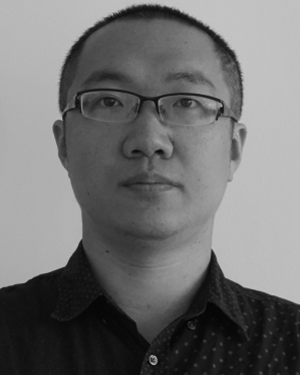}}]{Fei Chao}
(Member, IEEE) received the B.Sc. degree in mechanical engineering from the Fuzhou University, Fuzhou, China, in 2004, the M.Sc. degree with distinction in computer science from the University of Wales, Aberystwyth, U.K., in 2005, and the Ph.D. degree in robotics from the Aberystwyth University, Wales, U.K., in 2009.

He is currently an Associate Professor with the School of Informatics, Xiamen University, Xiamen, China. He has authored/co-authored more than 50 peer-reviewed journal and conference papers. His current research interests include developmental robotics, machine learning, and optimization algorithms.
\end{IEEEbiography}


\begin{IEEEbiography}[{\includegraphics[width=1in,height=1.25in,clip,keepaspectratio]{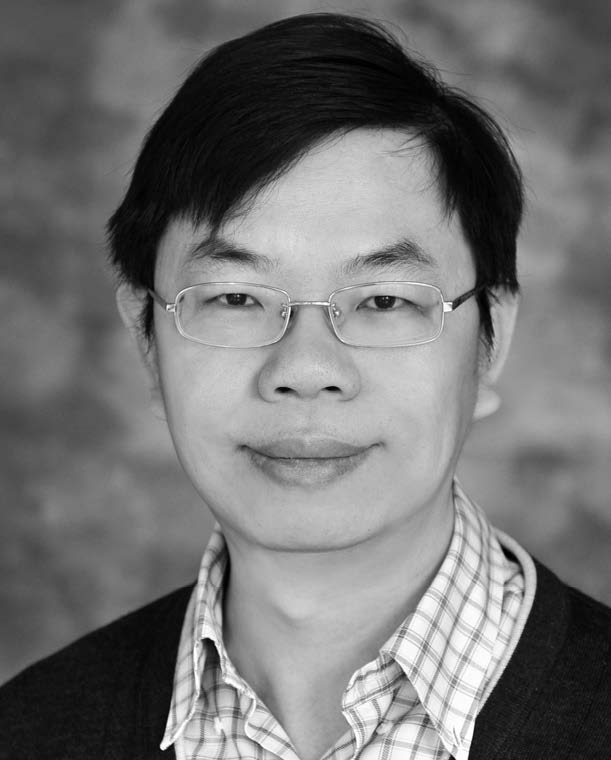}}]{Chia-Wen Lin} (Fellow, IEEE) received the Ph.D degree in electrical engineering from National Tsing Hua University (NTHU), Hsinchu, Taiwan, in 2000.
He is currently Professor with the Department of Electrical Engineering and the Institute of Communications Engineering, NTHU. He is also Deputy Director of the AI Research Center of NTHU. He was with the Department of Computer Science and Information Engineering, National Chung Cheng University, Taiwan, during 2000--2007. Prior to joining academia, he worked for the Information and Communications Research Laboratories, Industrial Technology Research Institute, Hsinchu, Taiwan, during 1992--2000. His research interests include image and video processing, computer vision, and video networking.

Dr. Lin served as  Distinguished Lecturer of IEEE Circuits and Systems Society from 2018 to 2019, a Steering Committee member of \textsc{IEEE Transactions on Multimedia} from 2014 to 2015, and the Chair of the Multimedia Systems and Applications Technical Committee of the IEEE Circuits and Systems Society from 2013 to 2015.  His articles received the Best Paper Award of IEEE VCIP 2015
and the Young Investigator Award of VCIP 2005. He received Outstanding Electrical Professor Award presented by Chinese Institute of Electrical Engineering in 2019, and Young
Investigator Award presented by Ministry of Science and Technology, Taiwan,
in 2006. He is also the Chair of the Steering Committee of IEEE ICME.  He has served as a Technical Program Co-Chair for IEEE ICME 2010, and a General Co-Chair for IEEE VCIP 2018, and a Technical Program Co-Chair for IEEE ICIP 2019. He has served as an Associate Editor of \textsc{IEEE Transactions on Image Processing}, \textsc{IEEE Transactions on Circuits and Systems for Video Technology}, \textsc{IEEE Transactions on Multimedia}, \textsc{IEEE Multimedia}, and \textit{Journal of Visual Communication and Image Representation}. 
\end{IEEEbiography}

\begin{IEEEbiography}[{\includegraphics[width=1in,height=1.25in,clip,keepaspectratio]{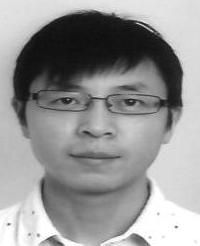}}]{Ling Shao}
(Fellow, IEEE) is the Chief Scientist of Terminus Group and the President of Terminus International. He was the founding CEO and Chief Scientist of the Inception Institute of Artificial Intelligence, Abu Dhabi, UAE. His research interests include computer vision, deep learning, medical imaging and vision and language. He is a fellow of the IEEE, the IAPR, the BCS and the IET.
\end{IEEEbiography}

\end{document}